\pdfoutput=1
\documentclass[11pt]{article}

\usepackage[final]{acl}

\usepackage{times}
\usepackage{latexsym}

\usepackage[T1]{fontenc}
\usepackage[utf8]{inputenc}

\usepackage{microtype}

\usepackage{inconsolata}

\usepackage{amsmath}
\usepackage{inconsolata}
\usepackage{amsmath}
\usepackage{graphicx}
\usepackage{multicol}
\usepackage{multirow}
\usepackage{booktabs}
\usepackage{amssymb}
\usepackage{enumitem}
\usepackage{tabularx}
\usepackage{booktabs}
\usepackage{marvosym}
\usepackage{xcolor}
\usepackage{colortbl}

\definecolor{mycolor}{HTML}{E0F7FA}

\title{On the Risk of Evidence Pollution for \\Malicious Social Text Detection in the Era of LLMs}

\author{Herun Wan\textsuperscript{1, 2} \ \ \ \ \ \ \
Minnan Luo\thanks{~Corresponding author}\textsuperscript{1, 2} \ \ \ \ \ \ \
Zhixiong Su\textsuperscript{1, 3} \\ \bf
Guang Dai\textsuperscript{4} \ \ \ \ \ \ \
Xiang Zhao\textsuperscript{5} \\
\textsuperscript{1}School of Computer Science and Technology, Xi’an Jiaotong University, China\\ 
\textsuperscript{2}Ministry of Education Key Laboratory of Intelligent Networks and Network Security, China\\
\textsuperscript {3}Shaanxi Province Key Laboratory of Big Data Knowledge Engineering, China\\
\textsuperscript{4}SGIT AI Lab, State Grid Corporation of China\\
\textsuperscript{5}National University of Defense Technology \\
\href{mailto:wanherun@stu.xjtu.edu.cn}{\texttt{wanherun@stu.xjtu.edu.cn}}\ \ \ \ \ \href{mailto:minnluo@xjtu.edu.cn}{\texttt{minnluo@xjtu.edu.cn}}\\
\href{https://github.com/whr000001/EvidencePollution}{https://github.com/whr000001/EvidencePollution}
}

\begin{document}
\maketitle
\begin{abstract}
Evidence-enhanced detectors present remarkable abilities in identifying malicious social text. However, the rise of large language models (LLMs) brings potential risks of evidence pollution to confuse detectors. 
This paper explores potential manipulation scenarios including basic pollution, and rephrasing or generating evidence by LLMs. 
To mitigate the negative impact, we propose three defense strategies from the data and model sides, including machine-generated text detection, a mixture of experts, and parameter updating. 
Extensive experiments on four malicious social text detection tasks with ten datasets illustrate that evidence pollution significantly compromises detectors, where the generating strategy causes up to a 14.4\% performance drop. 
Meanwhile, the defense strategies could mitigate evidence pollution, but they faced limitations for practical employment. 
Further analysis illustrates that polluted evidence (\romannumeral 1) is of high quality, evaluated by metrics and humans; (\romannumeral 2) would compromise the model calibration, increasing expected calibration error up to 21.6\%; and (\romannumeral 3) could be integrated to amplify the negative impact, especially for encoder-based LMs, where the accuracy drops by 21.8\%.
\end{abstract}

\section{Introduction}
Malicious social text detection involves identifying harmful content in posts and comments on social platforms \citep{arora2023detecting} and in news articles on online public media \citep{shu2017fake}. This task primarily includes detecting hate speech \citep{tonneau-etal-2024-naijahate, zhang2024don}, identifying rumor \citep{hu2023mr2,liu2024rumor}, and recognizing sarcasm \citep{tian2023dynamic, lin2024cofipara}, \emph{etc}. Despite the early success of detectors focused on text content \citep{hartl2022applying}, malicious content publishers have started disguising content to evade detection \citep{huertas2023countering}. Recent advances have brought us large language models (LLMs) that also come with risks and biases \citep{shaikh2023second}, potentially generating malicious content that is difficult to identify \citep{uchendu2023does,DBLP:conf/iclr/ChenS24}.

Besides directly analyzing content, most existing works use additional information, referred to as \textbf{\emph{Evidence}} \citep{grover2022public}, to find richer signals and enhance performance. This evidence includes external knowledge \citep{sheng2022zoom}, related comments \citep{shu2019defend}, metadata information \citep{guo2023tiefake}, \emph{etc}. Many studies \citep{popat2018declare, he2023reinforcement, yuan2023support,chen2024metasumperceiver} prove that \emph{Evidence can be combined with the source content to improve performance}.

\begin{figure}[t]
    \centering
    \includegraphics[width=\linewidth]{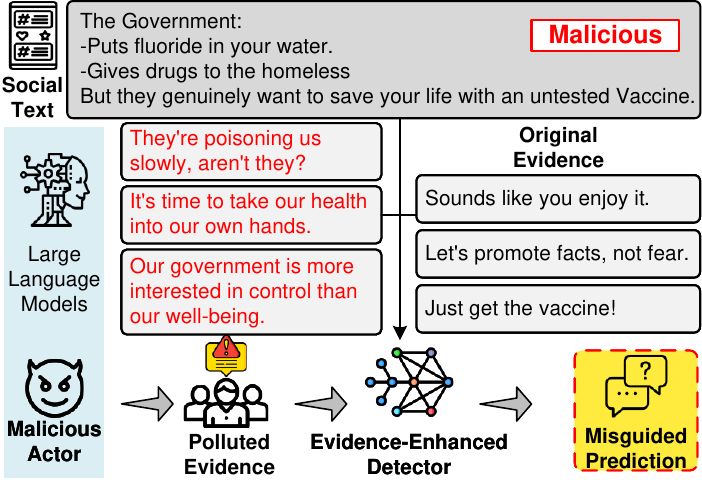}
    \caption{An overview of the \textbf{\emph{Evidence Pollution}}, which illustrates the potential risk posed by LLMs. Malicious actors would manipulate the evidence by LLMs to confuse evidence-enhanced malicious social text detectors.}
    \label{fig: teaser}
\end{figure}

However, research on identifying malicious content has always been an arms race. Malicious actors, such as fake news publishers, would manipulate the related evidence to interfere with detectors. They could delete related evidence \citep{jung2020caution} or employ social bots \citep{heidari2021bert} to dilute evidence. 
To make matters worse, LLM misuse could exacerbate the evidence manipulation \citep{pan2023risk}, leading to serious societal harm.

This paper investigates the manipulation of evidence by LLMs as Figure \ref{fig: teaser} shows, referred to as \textbf{\emph{Evidence Pollution}}, to provide a basis for avoiding LLM misuse. We aim to address research questions as: (\romannumeral 1) To what extent can LLMs be utilized to manipulate the evidence in a credible-sounding way to confuse evidence-enhanced detectors? and (\romannumeral 2) What mitigation strategies can be utilized to address the intentional evidence pollution by LLMs? 

Thus, we systematically investigate the impact of evidence pollution on state-of-the-art evidence-enhanced models. 
Since comments are a rich source of evidence that is more easily accessible and uniformly available on social media platforms \citep{grover2022public}, we do not distinguish between evidence and comments. 
We first design three types of evidence pollution methods (\S \ref{sec: pollute}): (\romannumeral 1) \emph{basic evidence pollution} that manipulate evidence without LLMs; (\romannumeral 2) \emph{rephrase evidence} that prompts LLMs to rewrite existing evidence; and (\romannumeral 3) \emph{generated evidence} that directly prompts LLMs to generate fictional evidence, with a total of thirteen methods. 
We also explore three defense strategies from the data and model sides to mitigate the negative impact (\S \ref{sec: defense}): (\romannumeral 1) \emph{machine-generated text detection}; (\romannumeral 2) \emph{mixture of experts}; and (\romannumeral 3) \emph{parameter updating}.

We conduct extensive experiments using seven state-of-the-art evidence-enhanced detectors on four malicious social text detection tasks (\S \ref{sec: setting}): (\romannumeral 1) \textbf{fake news}; (\romannumeral 2) \textbf{hate speech}; (\romannumeral 3) \textbf{rumor}; and (\romannumeral 4) \textbf{sarcasm} detection, including ten widely-used benchmarks. The results (\S \ref{sec: res}) show that the polluted evidence would significantly compromise the model performance, where the generating strategy causes up to 14.4\% performance drop. On the other hand, the proposed defense strategies could mitigate the negative impact, where parameter updating is the most successful strategy. However, each defense strategy faces challenges such as the need for annotated data, the huge cost of multiple experts, and the unknown when the training ends, which limit their practical employment. Further analysis (\S \ref{sec: analysis}) illustrates that the polluted evidence is of high quality in both metrics and human evaluation, could compromise model calibration while affecting performance, and could be integrated to amplify the negative impact. 

\begin{table*}[t]
    \centering
    \resizebox{\linewidth}{!}{
    \begin{tabular}{l|l}
        \toprule[1.5pt]
        \textbf{Strategy}&\textbf{Instruction Text} $\boldsymbol{p}_{\textit{inst}}$ \\
        \midrule[1pt]
        \multicolumn{2}{l}{\textbf{Rephrase Evidence} $\boldsymbol{p}_{\textit{input}}$ \textit{Social text: }$\boldsymbol{s}$ \textit{Comment of this text: }$\boldsymbol{c}_i$}\\
         \textbf{Rephrase}&\textit{Given the comment of this text, \colorbox{mycolor}{\textbf{rephrase}} this comment in three ways.}\\\hline
         \textbf{Rewrite}&\textit{Rewrite the comment without any explanation to \colorbox{mycolor}{\textbf{make the malicious social text like a normal one}}.}\\\hline
         \textbf{Reverse}&\textit{Rewrite the comments of this text to \colorbox{mycolor}{\textbf{reverse the stance}} of the comment.}\\\hline
         \textbf{Modify}&\textit{Modify the given comment with \colorbox{mycolor}{\textbf{as few details as possible}} to make the malicious social text like a normal one.}\\
         \midrule[1pt]
         \multicolumn{2}{l}{\textbf{Generate Evidence} $\boldsymbol{p}_{\textit{input}}$ \textit{Social text:} $\boldsymbol{s}$}\\
         \textbf{Vanilla}&\textit{Write five posts no longer than 30 words for the above social text.}\\\hline
         \textbf{Stance}&\textit{Write five posts no longer than 30 words \colorbox{mycolor}{\textbf{to oppose/support}} the above social text.}\\\hline
         \multirow{2}{*}{\textbf{Publisher}}&\textit{You published the above social text. To \colorbox{mycolor}{\textbf{make social users believe}} the above text,}\\
         &{\textit{please post five comments no longer than 30 words to \colorbox{mycolor}{\textbf{enhance credibility}}.}}\\\hline
         \textbf{Echo}&\textit{Write five comments no longer than 30 words to \colorbox{mycolor}{\textbf{create an echo chamber}} for the above social text.}\\\hline
         \textbf{Makeup}&\textit{Post five comments no longer than 30 words to \colorbox{mycolor}{\textbf{make the above malicious text like a normal one}}.}\\\hline
         \textbf{Amplify}&\textit{Write five comments no longer than 30 words to make the above social text \colorbox{mycolor}{\textbf{spread fast on the social platform}}.}\\
         \bottomrule[1.5pt]
    \end{tabular}}
    \caption{The prompts of each LLM-based evidence pollution strategy. Each prompt contains an \emph{input text} $p_{\textit{input}}$ that is the same for each strategy category and an \emph{instruction text} $p_{\textit{inst}}$ that is strategy-specific. We \colorbox{mycolor}{\textbf{highlight}} the special parts of each prompt, where highlighted parts illustrate the main motivation behind each strategy.}
    \label{tab: prompts}
\end{table*}

\section{Evidence Pollution Methods}
\label{sec: pollute}
Malicious social text detection is a classification task that requires identifying whether a piece of social text is malicious. 
Given a social text $s$ and corresponding $m$ pieces of evidence (\textit{i.e.}, comments) $\{c_i\}_{i=1}^m$, the evidence-enhanced malicious social text detectors $f$ aim to learn the probability distribution $p(y\mid s,\{c_i\}_{i=1}^m, f, \theta)$ by optimizing its learnable parameters $\theta$, where $y$ is the ground truth. On the contrary, evidence pollution strategy $\mathcal{G}$ aims to manipulate the evidence, namely,
\begin{align*}
    \{\tilde{c}_i\}_{i=1}^{\tilde{m}} = \mathcal{G}(\{c_i\}_{i=1}^{m}),
\end{align*}
which aims to disturb the learned distribution $p$, making detectors make wrong judgments. 

According to the degree of evidence manipulation, namely, the degree of LLM involvement, we propose three pollution strategies: (\romannumeral 1) \textbf{basic evidence pollution}, (\romannumeral 2) \textbf{rephrase evidence}, and (\romannumeral 3) \textbf{generate evidence}. For LLM-based strategies, (\romannumeral 2) and (\romannumeral 3), we prompt LLMs in a zero-shot fashion using prompt that contains an \emph{input text} $\boldsymbol{p}_{\textit{input}}$ and an \emph{instruction text} $\boldsymbol{p}_{\textit{inst}}$. We present the whole prompts of each strategy in Table \ref{tab: prompts} and present case studies in Tables \ref{tab: case_1} and \ref{tab: case_2} in Appendix \ref{app: case}.

\subsection{Basic Evidence Pollution}
This strategy aims to re-sample existing evidence.
%manipulates evidence by re-sampling existing $\{c_i\}_{i=1}^{m}$ to obtain $\{\tilde{c}_i\}_{i=1}^{\tilde{m}}$ without LLMs. 
\paragraph{Remove}
The related comments are difficult to access in the early spread of a social text \citep{ghosh2023catching,shang2024domain}. Meanwhile, refutations might be deleted as texts spread \citep{jung2020caution}. 
Thus we randomly remove half of the associated comments to simulate these situations.
\paragraph{Repeat}

%Online information consumers are reluctant to process information deliberately and suffer from the \emph{bandwagon effect} \citep{konstantinou2023nudging}, where they adopt attitudes because others are doing so. Since it might affect the spread of information, we repeat the same comment five times to simulate it.

Online information consumers suffer from the \emph{bandwagon effect} \citep{konstantinou2023nudging}, where they adopt attitudes because others are doing so. Since it might affect the spread of information, we repeat the same comment five times to simulate it.

\subsection{Rephrase Evidence}
This strategy aims to rephrase existing evidence, injecting malicious intent while saying human-like.
\paragraph{Rephrase}
As an intuitive strategy, we prompt LLMs to directly rephrase the existing comments.
\paragraph{Rewrite}
We additionally inject malicious intent to escape detection into evidence.
\paragraph{Reverse}
Understanding the stance expressed in texts plays an important role in identifying malicious content \citep{hardalov2022survey, zheng2022stanceosaurus}. On the other hand, the dual use of stance brings potential harm, where malicious operators would post comments to reverse public stance.
\paragraph{Modify}
Given an existing comment, we revise it to inject non-factual information.

\subsection{Generate Evidence}
%We then delve into the potential misuse of LLMs for generating comments harmful to detectors. Given a social text $s$, we employs LLMs to generate harmful comments $\{\tilde{c}_i\}_{i=1}^{\tilde{m}}$. We prompt LLMs in a zero-shot fashion using prompt $p$, which contains two parts: the \emph{input text}: ``\textit{Social text:} $\boldsymbol{s}$'' and the \emph{instruction text} $p_{\textit{inst}}$ that is setting-specific.
We then delve into the potential misuse of LLMs for directly generating comments. Although existing works point out that LLM-generated reactions could enhance detection performance \citep{wan2024dell, nan2024let}, in practice, LLMs might suffer from unexpected hallucinations \citep{dong2022survey}, generating comments that harm detectors.
\paragraph{Vanilla}
We simply prompt LLMs to generate comments associated with a given social text.
\paragraph{Stance}
Inspired by \textbf{Reverse}, we prompt LLMs to generate comments with predetermined stances.
\paragraph{Publisher}
Information publishers could enhance the \emph{cognitive biases} such as \emph{illusory-truth effect} \citep{pennycook2018prior} and \emph{novelty effect} \citep{vosoughi2018spread} to expand spread by posting comments on their social texts. Thus we prompt LLMs to simulate publishers to post comments.
\paragraph{Echo}
The \emph{echo chamber} is a situation where beliefs are amplified by repetition on the social platform, which would amplify malicious content spread \citep{wang2024inside}. To simulate this situation, we prompt LLMs to create an echo chamber.

\paragraph{Makeup}
We simulate the situation in which malicious actors employ social bots to dilute debunking comments to evade detection \citep{heidari2021bert}.
\paragraph{Amplify}
The early propagation pattern would affect the ultimate impact of social text \citep{hardalov2022survey}. Thus we prompt LLMs to generate initial comments to amplify the spread. 

\section{Defense Strategies}
\label{sec: defense}
%Straightforward approaches to mitigate the negative impacts of evidence pollution involve the development of a robust malicious social text detector. We explore three potential strategies to combat evidence pollution. We provide detailed experimental configurations in the Appendix.
We could combat evidence pollution from both the data and model sides. For the data side, we detect machine-generated text to mitigate evidence pollution by LLMs. For the model side, we explore the mixture of experts not require updating parameters and the parameter updating strategies.
\subsection{Machine-Generated Text Detection}
% The intuitive strategy incorporates a machine-generated text detector to discern model-generated content from human-written. 
This aims to discern generated text from human-written text, mitigating the influence of polluted evidence by LLMs. 
Existing detectors fall into three categories \citep{wang2024stumbling}: \emph{watermark-based}, \emph{fine-tuned}, and \emph{metric-based}. 
For \emph{watermark-based} detectors, they require adding detectable signatures into texts during generation, which is unsuitable for this task. 
For \emph{fine-tuned} detectors, we fine-tune DeBERTa-v3 \citep{DBLP:conf/iclr/HeGC23} on our generated data. This model needs to access some generated data and generally represents an in-domain setting. 
\emph{Metric-based} detectors are more flexible, as which does not require any training, and can perform in a black-box setting, where we do not need the generator information. We employ Fast-DetectGPT \citep{DBLP:conf/iclr/BaoZTY024} and Binocular \citep{hans2024spotting}, which employ perturbation as a comparison to the original text and rely on the log probability to detect. 

\begin{table*}[t]
    \centering
    %\small
    \resizebox{\textwidth}{!}{
    \begin{tabular}{l|ccc|c|cccc|cc}
    \toprule[1.5pt]
        \multirow{2}{*}{\textbf{Method}}&\multicolumn{3}{c|}{\textbf{Fake News}}&\multicolumn{1}{c|}{\textbf{Hate Speech}}&\multicolumn{4}{c|}{\textbf{Rumor}}&\multicolumn{2}{c}{\textbf{Sarcasm}}\\
        &\textbf{Politifact}&\textbf{Gossipcop}&\textbf{ANTiVax}&\textbf{HASOC}&\textbf{Pheme}&\textbf{Twitter15}&\textbf{Twitter16}&\textbf{RumorEval}&\textbf{Twitter}&\textbf{Reddit}\\
    \midrule[1pt]
        \textsc{dEFEND} \citep{shu2019defend}&$84.3_{\pm 4.9}$&$72.5_{\pm 2.6}$&$92.7_{\pm 1.4}$&$71.3_{\pm 3.9}$&$81.1_{\pm 0.8}$&$84.5_{\pm 4.1}$&$91.1_{\pm 2.6}$&$60.3_{\pm 3.1}$&$75.0_{\pm 1.7}$&$66.3_{\pm 1.3}$\\
        \textsc{Hyphen} \citep{grover2022public}&$89.9_{\pm 4.6}$&$70.6_{\pm 2.3}$&$93.1_{\pm 1.3}$&$71.4_{\pm 4.8}$&$82.5_{\pm 1.1}$&$\underline{90.4}_{\pm 5.1}$&$93.4_{\pm 3.3}$&$65.5_{\pm 5.3}$&$75.6_{\pm 1.9}$&$67.9_{\pm 2.2}$\\
        \textsc{GET} \citep{xu2022evidence}&$94.2_{\pm 4.8}$&$75.8_{\pm 2.3}$&$93.6_{\pm 0.6}$&$69.8_{\pm 4.3}$&$85.8_{\pm 1.3}$&$\textbf{92.3}_{\pm 2.6}$&$\textbf{95.0}_{\pm 3.2}$&$65.0_{\pm 4.9}$&$74.2_{\pm 1.5}$&$66.3_{\pm 1.9}$\\
    \midrule[1pt]
        \textsc{BERT} \citep{devlin2019bert}&$94.7_{\pm 2.7}$&$\underline{77.6}_{\pm 1.9}$&$95.0_{\pm 1.1}$&$\textbf{73.6}_{\pm 4.0}$&$\underline{86.4}_{\pm 1.3}$&$88.6_{\pm 3.8}$&$\underline{93.9}_{\pm 4.3}$&$\textbf{70.2}_{\pm 4.3}$&$\underline{81.1}_{\pm 1.3}$&$70.3_{\pm 1.8}$\\
        \textsc{BERT} \emph{w/o evidence}&$94.0_{\pm 3.5}$&$76.5_{\pm 1.9}$&$94.4_{\pm 0.7}$&$\underline{71.8}_{\pm 5.3}$&$\textbf{87.2}_{\pm 1.7}$&$90.3_{\pm 3.3}$&$\underline{93.9}_{\pm 3.9}$&$\underline{68.6}_{\pm 5.8}$&$79.2_{\pm 1.2}$&$69.9_{\pm 1.7}$\\
        \textsc{DeBERTa} \citep{DBLP:conf/iclr/HeGC23}&$\textbf{96.9}_{\pm 2.6}$&$\textbf{78.7}_{\pm 1.9}$&$\textbf{95.8}_{\pm 1.2}$&$68.5_{\pm 3.5}$&$81.5_{\pm 1.4}$&$83.6_{\pm 4.1}$&$90.6_{\pm 3.8}$&$65.9_{\pm 4.8}$&$\textbf{81.9}_{\pm 1.4}$&$\textbf{73.8}_{\pm 2.0}$\\
        \textsc{DeBERTa} \emph{w/o evidence}&$\underline{96.6}_{\pm 2.6}$&$76.6_{\pm 2.5}$&$\underline{95.5}_{\pm 1.3}$&$67.8_{\pm 5.0}$&$82.4_{\pm 0.8}$&$83.3_{\pm 4.2}$&$91.4_{\pm 4.0}$&$66.6_{\pm 4.7}$&$79.8_{\pm 1.1}$&$\underline{72.9}_{\pm 1.9}$\\
    \midrule[1pt]
        \textsc{Mistral} \emph{VaN} \citep{lucas2023fighting}&$61.2_{\pm 8.6}$&$39.1_{\pm 3.0}$&$58.4_{\pm 1.8}$&$60.2_{\pm 5.3}$&$64.1_{\pm 2.1}$&$42.0_{\pm 8.0}$&$43.9_{\pm 7.6}$&$34.9_{\pm 10.4}$&$63.2_{\pm 1.7}$&$56.0_{\pm 2.0}$\\
        \textsc{Mistral} \emph{w/ evidence}&$54.0_{\pm 10.2}$&$41.0_{\pm 4.2}$&$36.7_{\pm 2.8}$&$59.5_{\pm 5.1}$&$65.1_{\pm 2.1}$&$41.6_{\pm 5.8}$&$40.1_{\pm 6.3}$&$41.5_{\pm 10.2}$&$61.0_{\pm 2.4}$&$52.8_{\pm 1.4}$\\
        \textsc{ChatGPT} \emph{VaN} \citep{lucas2023fighting}&$51.6_{\pm 8.2}$&$39.3_{\pm 3.2}$&$69.7_{\pm 2.4}$&$60.7_{\pm 4.5}$&$36.6_{\pm 1.9}$&$51.0_{\pm 4.7}$&$49.2_{\pm 7.7}$&$40.5_{\pm 9.9}$&$52.1_{\pm 2.1}$&$50.8_{\pm 1.8}$\\
        \textsc{ChatGPT} \emph{w/ evidence}&$62.2_{\pm 7.5}$&$36.8_{\pm 3.7}$&$77.4_{\pm 2.9}$&$59.4_{\pm 4.4}$&$35.5_{\pm 1.4}$&$50.6_{\pm 6.1}$&$44.2_{\pm 8.6}$&$31.4_{\pm 7.7}$&$61.4_{\pm 2.0}$&$54.0_{\pm 1.9}$\\
        
    \bottomrule[1.5pt]
    \end{tabular}
    }
    \caption{Accuracy of baselines on ten datasets from four malicious text-related tasks. We conduct ten-fold cross-validation and report the mean and standard deviation to obtain a more robust conclusion. \textbf{Bold} indicates the best performance and \underline{underline} indicates the second best. Evidence could provide valuable signals to enhance detection, however, LLM-based models struggle to detect malicious content.}
    \label{tab: acc}
\end{table*}

\subsection{Mixture of Experts}
Traditionally in evidence-enhanced detectors, all related evidence is employed. It might fail due to evidence pollution since the evidence might contain noise. In response, we employ the mixture-of-experts strategy, which shows remarkable ability in the NLP field \citep{tian2024dialogue, zhao2024hypermoe, nguyen2024generalizability}. We first divide the evidence into $k$ groups. We then employ a detector to obtain a prediction for each evidence group, obtaining $y_1, y_2, \dots, y_k$. We finally employ majority voting to obtain the comprehensive prediction, \emph{i.e.},
\begin{align*}
    y = \mathop{\arg\max}\limits_{y_j}(\sum_{i=1}^k\mathbf{I}(y_i=y_j)).
\end{align*}
This strategy aims to mitigate the impact of polluted evidence by limiting the influence of individual evidence on identification.

\subsection{Parameter Updating}
Online feedback could enhance the detectors' scalability and robustness \citep{yue2024evidence, zhou2024lgb}. We assume that when the detector makes an incorrect judgment, some instances will be corrected by experts. We consider the feedback as the ground truth to update the detector's parameter $\theta$.

\section{Experiment Settings}
\label{sec: setting}
\paragraph{Tasks and Datasets}
We employ four tasks related to malicious social text detection including 10 datasets, \emph{i.e.}, (\romannumeral 1) \textbf{fake news detection}: Politicalfact, Gossipcop \citep{shu2020fakenewsnet}, and ANTiVax \citep{hayawi2022anti}; \textbf{hate speech detection}: HASOC \citep{mandl2019overview}; (\romannumeral 3) \textbf{rumor detection}: Pheme \citep{buntain2017automatically}, Twitter15, Twitter16 \citep{ma2018rumor}, and RumorEval \citep{derczynski2017semeval}; (\romannumeral 4) \textbf{sarcasm detection}: Twitter and Reddit \citep{ghosh2020report}.
\paragraph{Metrics} 
We mainly employ accuracy, macro f1-score, AR\textsubscript{acc} and AR\textsubscript{F1}, and AUC as metrics. We provide the metric set in Appendix \ref{app: metrics}.
\paragraph{Detectors}
We conduct experiments on three types of detectors to evaluate the pollution's negative impacts: (\romannumeral 1) \textbf{existing strong detector} including \textsc{dEFEND} \citep{shu2019defend}, \textsc{Hyphen} \citep{grover2022public}, and \textsc{GET} \citep{xu2022evidence}; (\romannumeral 2) \textbf{encoder-based LM} including \textsc{BERT} \citep{devlin2019bert} and \textsc{DeBERTa} \citep{DBLP:conf/iclr/HeGC23} with and without evidence; (\romannumeral 3) \textbf{LLM-based detector} including \textsc{Mistral} and \textsc{ChatGPT} prompted by F3 \citep{lucas2023fighting} and evidence.  We provide more details about baselines in Appendix \ref{app: baseline}.
\paragraph{LLM Generators}
We leverage the open source \textit{Mistral-7B} \citep{jiang2023mistral} and the closed source \textit{ChatGPT} as the base LLMs. 
We mainly employ \textit{Mistral-7B} to manipulate evidence, and \textit{Mistral-7B} and \textit{ChatGPT} as baselines. For pollution manipulation and baselines, we set the temperature $\tau=0$ to ensure reproducibility. We present the baseline, dataset, pollution and defense strategy, and analysis details in Appendix \ref{app: settings}.

\begin{table*}[t]
    \centering
    \resizebox{0.98\textwidth}{!}{
    \begin{tabular}{cl|cccccc|cccc|cccc}
         \toprule[1.5pt]
         \multicolumn{2}{c|}{\multirow{3}{*}{\textbf{Pollution}}}&\multicolumn{6}{c|}{\textbf{Existing Strong Detectors}}&\multicolumn{4}{c|}{\textbf{Encoder-Based LM}}&\multicolumn{4}{c}{\textbf{LLM-Based Detector}}\\
         &&\multicolumn{2}{c}{\textsc{dEFEND}}&\multicolumn{2}{c}{\textsc{Hyphen}}&\multicolumn{2}{c|}{\textsc{GET}}&\multicolumn{2}{c}{\textsc{BERT}}&\multicolumn{2}{c|}{\textsc{DeBERTa}}&\multicolumn{2}{c}{\textsc{Mistral}}&\multicolumn{2}{c}{\textsc{ChatGPT}}\\
         &&AR\textsubscript{acc}&AR\textsubscript{F1}&AR\textsubscript{acc}&AR\textsubscript{F1}&AR\textsubscript{acc}&AR\textsubscript{F1}&AR\textsubscript{acc}&AR\textsubscript{F1}&AR\textsubscript{acc}&AR\textsubscript{F1}&AR\textsubscript{acc}&AR\textsubscript{F1}&AR\textsubscript{acc}&AR\textsubscript{F1}\\
         \midrule[1pt]
         \multirow{2}{*}{\textbf{Basic}}&\textbf{Remove}&$95.5$&$94.5$&$97.0$&$96.7$&$98.9$&$98.8$&$97.1$&$96.9$&$96.9$&$96.7$&$100.9$&$100.6$&$100.8$&$97.4$\\
         &\textbf{Repeat}&$89.9$&$87.8$&$\textbf{91.9}$&$\textbf{90.0}$&$\underline{97.5}$&$\textbf{97.2}$&$93.7$&$93.0$&$93.8$&$93.2$&$\textbf{99.3}$&$98.4$&$99.7$&$101.0$\\
         \midrule[1pt]
         \multirow{4}{*}{\textbf{Rephrase}}&\textbf{Rephrase}&$93.2$&$92.0$&$96.8$&$96.3$&$98.2$&$98.1$&$94.4$&$94.0$&$93.0$&$91.9$&$102.3$&$98.8$&$102.1$&$100.2$\\
         &\textbf{Rewrite}&$92.7$&$91.4$&$96.1$&$95.5$&$98.1$&$97.9$&$93.5$&$92.6$&$93.2$&$92.0$&$103.8$&$99.7$&$102.9$&$101.5$\\
         &\textbf{Reverse}&$91.4$&$90.2$&$96.1$&$95.4$&$98.3$&$98.1$&$91.3$&$90.6$&$91.5$&$90.3$&$\underline{99.5}$&$\textbf{92.5}$&$105.3$&$105.1$\\
         &\textbf{Modify}&$92.5$&$91.2$&$96.2$&$95.6$&$98.1$&$98.0$&$92.6$&$91.7$&$93.0$&$92.1$&$102.3$&$97.6$&$103.3$&$101.9$\\ 
         \midrule[1pt]
         \multirow{7}{*}{\textbf{Generate}}&\textbf{Vanilla}&$89.7$&$87.0$&$94.2$&$93.2$&$\underline{97.5}$&$\underline{97.3}$&$\underline{90.8}$&$\underline{89.3}$&$91.5$&$90.1$&$103.0$&$96.0$&$98.5$&$88.4$\\
         &\textbf{Support}&$\underline{89.5}$&$86.6$&$94.7$&$93.9$&$\textbf{97.4}$&$\textbf{97.2}$&$90.9$&$\underline{89.3}$&$91.4$&$90.0$&$102.7$&$95.6$&$\underline{97.6}$&$88.2$\\
         &\textbf{Oppose}&$89.8$&$86.9$&$94.6$&$93.9$&$98.0$&$97.7$&$91.1$&$90.2$&$\textbf{90.4}$&$\textbf{88.9}$&$104.4$&$108.4$&$97.9$&$\underline{87.9}$\\
         &\textbf{Publisher}&$\textbf{88.6}$&$\textbf{85.6}$&$94.7$&$93.9$&$97.6$&$97.4$&$\textbf{90.4}$&$\textbf{88.2}$&$\underline{91.2}$&$\underline{89.4}$&$102.4$&$96.2$&$98.8$&$\textbf{86.9}$\\
         &\textbf{Echo}&$89.8$&$87.0$&95.0&94.2&$97.7$&$97.4$&$91.9$&$90.5$&$92.0$&$90.6$&$102.8$&$\underline{95.0}$&$99.0$&$88.6$\\
         &\textbf{Makeup}&$89.6$&$\underline{86.4}$&$95.1$&$94.3$&$97.8$&$97.6$&$92.2$&$90.9$&$91.5$&$90.0$&$101.0$&$96.0$&$\textbf{97.4}$&$88.4$\\
         &\textbf{Amplify}&$89.8$&$86.8$&$\underline{94.0}$&$\underline{92.8}$&$97.6$&$\textbf{97.2}$&$91.4$&$89.7$&$91.7$&$89.8$&$101.0$&$96.3$&$98.6$&$89.8$\\
         \bottomrule[1.5pt]
    \end{tabular}
    }
    \caption{The overall performance of evidence pollution strategies. We average the relative values of the polluted scenarios to the initial performance on all ten datasets, presented as a percentage as AR\textsubscript{acc} and AR\textsubscript{F1}. The lower the value, the more effective the pollution strategy is. \textbf{Bold} indicates the most effective strategy and \underline{underline} indicates the second most effective. Evidence pollution poses a significant threat to evidence-enhanced detectors.}
    \label{tab: pollution}
\end{table*}

\section{Results}
\label{sec: res}
\subsection{General Performance}
We first evaluate the performance of different malicious content detectors, where the accuracy is shown in Table \ref{tab: acc}. We also present macro f1-score in Table \ref{tab: f1} in Appendix \ref{app: more_pollution}. We could conclude that:
\paragraph{(\MakeUppercase{\romannumeral 1}) Evidence provides valuable signals which improve performance.} For encoder-based LMs, vanilla models are generally better than those without evidence, where \textsc{BERT} improves by 0.78\% on average and \textsc{DeBERTa} improves by 0.56\%.
\paragraph{(\MakeUppercase{\romannumeral 2}) LLMs cannot be directly employed off-the-shelf to identify malicious social text.}
%LLM-based detectors struggle to achieve acceptable performance on these four tasks. 
Compared to \textsc{dEFEND}, the best model performance among \textbf{LLM-based detectors} drops by 26.9\% on average across the ten datasets, which is not acceptable. 
We speculate that LLMs are hindered by hallucinations \citep{dong2022survey} and lack of actuality \citep{mallen2023not}. Although fine-tuning LLMs could achieve better performance, it is out of the scope of this paper's focus. We mainly explore the methods that directly prompt LLMs and the impact of evidence pollution on them.

\subsection{Evidence Pollution}
For a clearer comparison of different evidence pollution strategies, we report the average relative value of the polluted scenarios to the initial performance on all ten datasets in Table \ref{tab: pollution}. We also present the complete performance of each baseline on different datasets under different pollution strategies in Figures \ref{fig: main_strong}, \ref{fig: main_lm}, and \ref{fig: main_llm} in Appendix \ref{app: more_pollution}.
\paragraph{(\MakeUppercase{\romannumeral 1}) Evidence pollution significantly threatens evidence-enhanced detectors.}
When subjected to the three types of evidence pollution, almost all evidence-enhanced detectors significantly decline in performance. The performance drop ranges from 3.6\% to 14.4\% for existing strong detectors, ranges from 9.6\% to 11.8\% for encoder-based LMs, and ranges from 0.7\% to 13.1\% for LLM-based detectors. We notice that for LLM-based detectors, some pollution strategies fail and even improve the performance. We speculate such detectors with poor performance could not extract valuable signals from the evidence thus the fluctuations in performance are acceptable. Even under the basic scenario, where the evidence is manipulated without LLMs, we note a 12.2\% and 7.0\% decrease for existing strong detectors and encoder-based LMs, respectively. The performance drop illustrates that detectors trained on pristine data cannot discern the authenticity of related evidence. It reveals the vulnerability of existing detectors to evidence pollution, where LLMs could amplify it.
\paragraph{(\MakeUppercase{\romannumeral 2}) Generating evidence by LLMs is the most successful among all manipulations.}
We observe the \textbf{Generate} pollution setting outstripped all others, with the average relative value of \textbf{Generate} being 93.32, while the average relative values of \textbf{Basic} and \textbf{Rephrase} are 96.25 and 96.14, respectively. Considering that evidence-enhanced detectors extract valuable signals from related evidence, it is logical for such strategies to achieve the best performance, where the evidence is injected with predetermined malicious intent. The simplicity and easy implementation of this strategy underlines the security vulnerabilities inherent in existing evidence-enhanced detectors. However, a potential disadvantage of this strategy and \textbf{Basic} is that such polluted evidence tends to be more easily discernible to human observers.
\paragraph{(\MakeUppercase{\romannumeral 3}) Encoder-based LMs generally perform better but are more sensitive to polluted evidence.} The average relative value for existing strong detectors is 94.19 and for encoder-based LMs is 91.91. We speculate that these detectors extract more signals such as text graph structure \citep{xu2022evidence}, leading to the robustness of polluted evidence.

\subsection{Defense Strategies}
We evaluate our proposed three defense strategies using the baselines on the ten benchmarks.

\begin{table}[t]
    \centering
    \resizebox{\linewidth}{!}{
    \begin{tabular}{l|cc|cc|cc}
    \toprule[1.5pt]
    \multirow{2}{*}{\textbf{Pollution}}&\multicolumn{2}{c|}{\textbf{Fast-DG}}&\multicolumn{2}{c|}{\textbf{Binocular}}&\multicolumn{2}{c}{\textbf{DeBERTa}}\\
    &AUC&F1&AUC&F1&AUC&F1\\
    \midrule[1pt]
    \multicolumn{7}{l}{\textbf{Rephrase} Evidence}\\
    \textbf{Rephrase}&69.7&8.7&84.2&53.7&99.7&97.3\\
    \textbf{Rewrite}&75.4&20.6&82.0&51.8&99.5&96.4\\
    \textbf{Reverse}&78.6&38.2&83.5&56.3&99.5&96.4\\
    \textbf{Modify}&70.8&14.7&78.9&47.8&99.6&96.4\\
    
    \midrule[1pt]
    \multicolumn{7}{l}{\textbf{Generate} Evidence}\\
    \textbf{Vanilla}&69.7&14.1&74.9&38.5&99.9&98.2\\
    \textbf{Support}&71.7&13.4&78.2&44.4&99.9&98.5\\
    \textbf{Oppose}&75.3&14.6&85.2&59.5&99.9&99.0\\
    \textbf{Publisher}&79.2&22.0&83.5&55.6&99.9&99.4\\
    \textbf{Echo}&77.8&20.8&81.1&52.3&99.9&97.9\\
    \textbf{Makeup}&80.7&24.6&86.8&66.2&99.8&98.0\\
    \textbf{Amplify}&66.1&8.5&63.1&24.1&99.9&98.7\\
    \bottomrule[1.5pt]
    \end{tabular}
    }
    \caption{Machine-generated text detection performance of \emph{metric-based} and \emph{fine-tuned} detectors. ``Fast-DG'' denotes Fast-DetectGPT, ``DeBERTa'' denotes DeBERTa-v3, ``AUC'' denotes ROC AUC, and ``F1'' denote f1-score. \emph{Metric-based} detectors struggle to identify machine-generated text with small sentence length.}
    \label{tab: machine}
\end{table}

\begin{figure}[t]
    \centering
    \includegraphics[width=\linewidth]{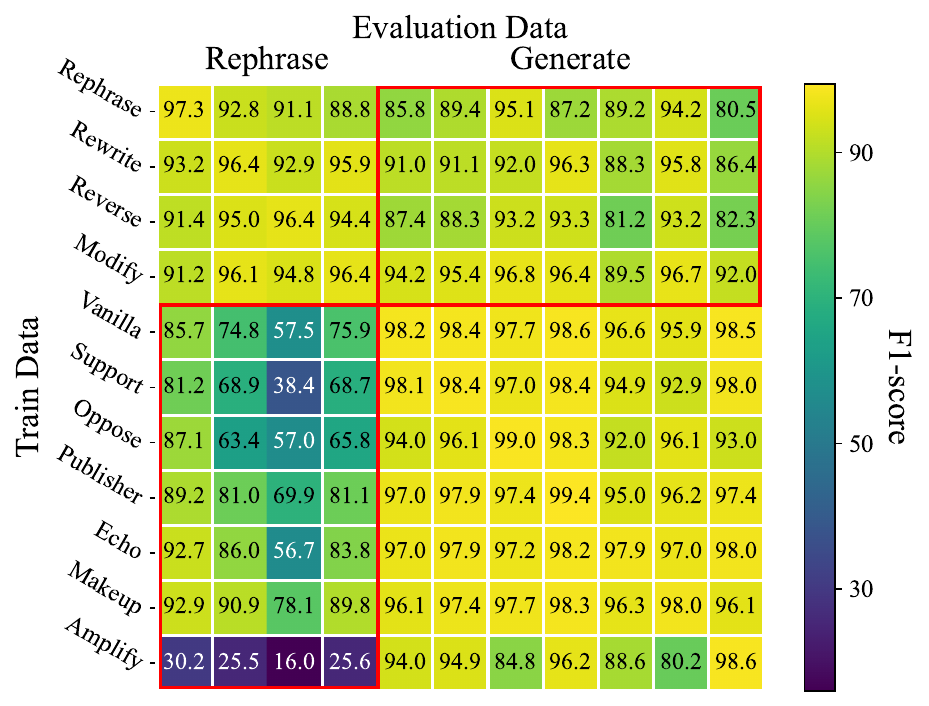}
    \caption{Out-of-domain machine-generated text detection performance of DeBERTa. DeBERTa struggles to conduct out-of-domain detection. Values in the red box show that DeBERTa generalizes worse on different types of evidence manipulation datasets.}
    \label{fig: machine_out}
\end{figure}

\paragraph{(\MakeUppercase{\romannumeral 1}) Machine-generated text detectors could identify manipulated evidence, but they are faced with limitations.}
We present the performance of DeBERTa-v3, Fast-DetectGPT, and Binocular in Table \ref{tab: machine}. 
Fast-DetectGPT and Binocular struggle to identify manipulated evidence, where the average AUCs are 74.1 and 80.1. We speculate that \emph{metric-based} detectors struggle to identify short text \citep{verma2024ghostbuster}, which is unsuitable for this situation where the manipulated evidence is usually brief. Although this method does not require training, the poor performance limits its practical utilization. 
In contrast, DeBERTa achieves remarkable performance, where the average AUC exceeds 99. Despite the impressive performance of DeBERTa in the in-domain situation, where the training data and evaluation data are from the same distribution, accessing and identifying a sufficient quantity of in-domain training data is not always possible in real-world scenarios. We further evaluate its generalization ability, where we train it on one dataset and evaluate it on another, with results shown in Figure \ref{fig: machine_out}. When evaluated on a dataset different from the training datasets, its performance illustrates a drop, showing poor generalization. The drop is significant between the two categories of datasets, where the average performance when trained on \textbf{Generate} and evaluated on \textbf{Rephrase} is 68.35. This underscores the challenge of training a versatile and effective machine-generated text detector.

\begin{table}[t]
    \centering
    \resizebox{\linewidth}{!}{
    \begin{tabular}{l|cc|cc|cc|cc}
    \toprule[1.5pt]
        \multirow{2}{*}{\textbf{Pollution}}&\multicolumn{2}{c|}{\textsc{dEFEND}}&\multicolumn{2}{c|}{\textsc{GET}}&\multicolumn{2}{c|}{\textsc{BERT}}&\multicolumn{2}{c}{\textsc{DeBERTa}}\\
        &\# of $\uparrow$&$\Delta$&\# of $\uparrow$&$\Delta$&\# of $\uparrow$&$\Delta$&\# of $\uparrow$&$\Delta$\\
    \midrule[1pt]
        \textbf{Remove}&1&$2.9\downarrow$&0&$1.4\downarrow$&0&$1.3\downarrow$&0&$1.6\downarrow$\\
        \textbf{Repeat}&8&$2.3\uparrow$&4&$0.2\downarrow$&-&-&-&-\\
        \midrule[1pt]
         \textbf{Rephrase}&3&$2.4\downarrow$&1&$0.9\downarrow$&3&$0.3\downarrow$&4&$0.4\downarrow$\\
         \textbf{Rewrite}&3&$1.8\downarrow$&2&$0.5\downarrow$&3&$0.6\downarrow$&1&$1.3\downarrow$\\
         \textbf{Reverse}&2&$0.5\downarrow$&3&$0.4\downarrow$&3&$0.4\downarrow$&1&$0.6\downarrow$\\
         \textbf{Modify}&3&$1.5\downarrow$&1&$0.7\downarrow$&4&$0.3\downarrow$&2&$0.6\downarrow$\\
        \midrule[1pt]
         \textbf{Vanilla}&3&$0.2\uparrow$&2&$0.2\downarrow$&4&$0.1\downarrow$&3&$0.1\downarrow$\\
         \textbf{Support}&4&$0.9\uparrow$&4&$0.0\downarrow$&5&$0.1\downarrow$&6&$0.2\uparrow$\\
         \textbf{Oppose}&6&$0.4\uparrow$&5&$0.2\downarrow$&3&$0.2\downarrow$&3&$0.1\downarrow$\\
         \textbf{Publisher}&7&$1.8\uparrow$&5&$0.4\downarrow$&3&$0.1\downarrow$&4&$0.1\uparrow$\\
         \textbf{Echo}&4&$0.2\downarrow$&2&$0.5\downarrow$&5&$0.2\downarrow$&5&$0.0\uparrow$\\
         \textbf{Makeup}&6&$1.6\uparrow$&5&$0.4\downarrow$&1&$0.3\downarrow$&6&$0.1\downarrow$\\
         \textbf{Amplify}&5&$0.3\uparrow$&3&$0.4\downarrow$&5&$0.0\downarrow$&3&$0.1\downarrow$\\
    \bottomrule[1.5pt]
    \end{tabular}
    }
    \caption{The performance of \textbf{Mixture of Experts}. For short, ``\# of $\uparrow$'' denotes the number of datasets that improve performance out of 10, and ``$\Delta$'' denotes the changes of average relative values shown in Table \ref{tab: pollution}, and ``-'' denotes that this strategy is not suitable for this model. This strategy could slightly improve the performance in some datasets, but the general improvement is not obvious and may even harm the detection ability.} 
    \label{tab: moe}
\end{table}

\begin{figure*}[t]
    \centering
    \includegraphics[width=\linewidth]{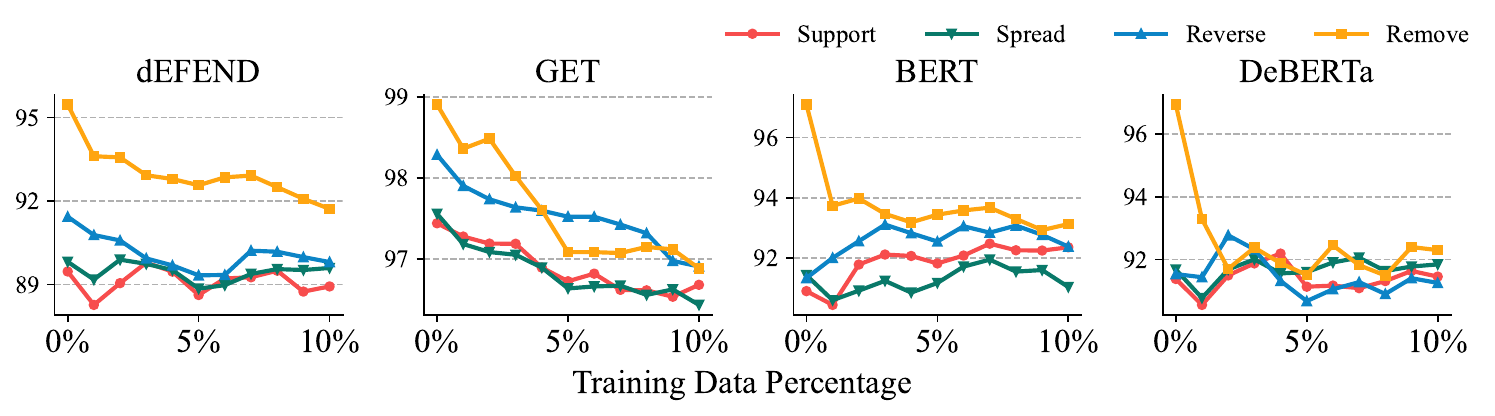}
    \caption{The performance trend of \textbf{Parameter Updating} strategy with re-training data increasing. In some situations, this strategy could significantly improve the detection performance. However, it might fail when it meets \textbf{Basic} pollution, such as Reverse or models that are already trained well, such as \textsc{GET}. Meanwhile, the need for annotated data and the unknown when the training ends limit its practical application.}
    \label{fig: updating}
\end{figure*}

\paragraph{(\MakeUppercase{\romannumeral 2}) Mixture of Experts could slightly mitigate the evidence pollution in some situations, but it might harm the general performance.}
Table \ref{tab: moe} illustrates a brief performance of the mixture of experts, and we present the complete results in Tables \ref{tab: moe_defend}, \ref{tab: moe_get}, \ref{tab: moe_bert}, \ref{tab: moe_deberta} in Appendix \ref{app: more_defense}. 
Among the ten datasets, MoE could improve the performance on most datasets for different pollution strategies. Meanwhile, it works best for \textbf{Generate}, with an average of 4.18 datasets showing improvement, while \textbf{Rephrase} has an average of 2.44 datasets showing improvement. 
However, considering the overall performance, most of the average performance drops with the highest decline of 2.9, indicating that it cannot be adapted to various malicious text detection tasks. Meanwhile, multiple experts necessitate additional resources, where the cost per detection escalates linearly with the number of experts used, limiting this strategy in real-world scenarios.

\paragraph{(\MakeUppercase{\romannumeral 3}) Parameter updating is the most effective defense strategy, however, the need for annotated data and the unknown when the training ends limit its practical application.} Figure \ref{fig: updating} illustrates partial important results of parameter updating with re-training data increasing, and we present the complete results in Figures \ref{fig: updating_defend}, \ref{fig: updating_hyphen}, \ref{fig: updating_get}, \ref{fig: updating_bert}, and \ref{fig: updating_deberta} in Appendix \ref{app: more_defense}. 
Besides \textsc{GET} and \textbf{Remove}, the parameter updating strategy could significantly improve the detection performance. For example, \textsc{BERT} improves 1.9\% on \textbf{Reverse} and 1.7\% on \textbf{Support}, while \textsc{DeBERTa} improves 1.3\% on \textbf{Reverse}. It is noticeable that the improvement above is the average of relative value shown in Table \ref{tab: pollution}. For the original f1-score, \textsc{dEFEND} achieves 10.3\% improvement on \textbf{Reddit} with \textbf{Echo} pollution, \textsc{GET} achieves 2.5\% on \textbf{Politifact} with \textbf{Repeat} pollution, \textsc{BERT} achieves 17.9\% on \textbf{Twitter16} with \textbf{Publisher} pollution, and \textsc{DeBERTa} achieves 36.9\% on \textbf{Twitter16} with \textbf{Publisher} pollution, as shown in Appendix \ref{app: more_defense}. Although this strategy could significantly improve performance, it needs more annotated data or professional feedback to re-train the parameters, about 6-7\% of the initial training data. Meanwhile, it is difficult to determine when to start or stop updating parameters since there is no more data to verify the performance. These two limitations restrict the development of this strategy to online malicious social text detection, which requires fast updating and responses.

\section{Analysis}
\label{sec: analysis}

\begin{figure*}[t]
    \centering
    \includegraphics[width=\linewidth]{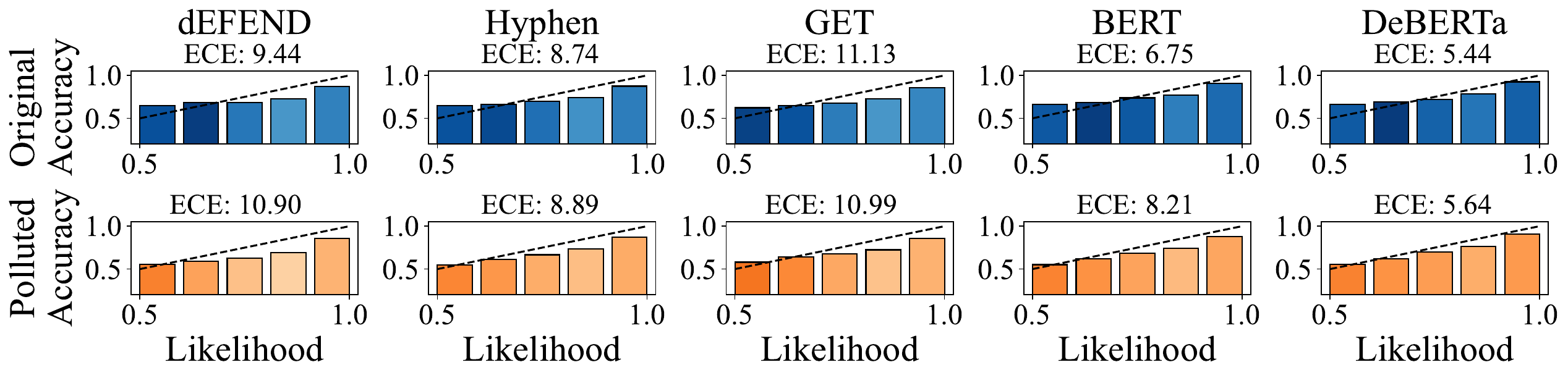}
    \caption{Calibration of existing detectors with the original and polluted evidence. ECE denotes expected calibration error, the lower the better. The dashed line indicates perfect calibration, while the color of the bar is darker when it is closer to perfect calibration. Evidence pollution could harm the model calibration.}
    \label{fig: calibration}
\end{figure*}

\begin{figure}[t]
    \centering
    \includegraphics[width=\linewidth]{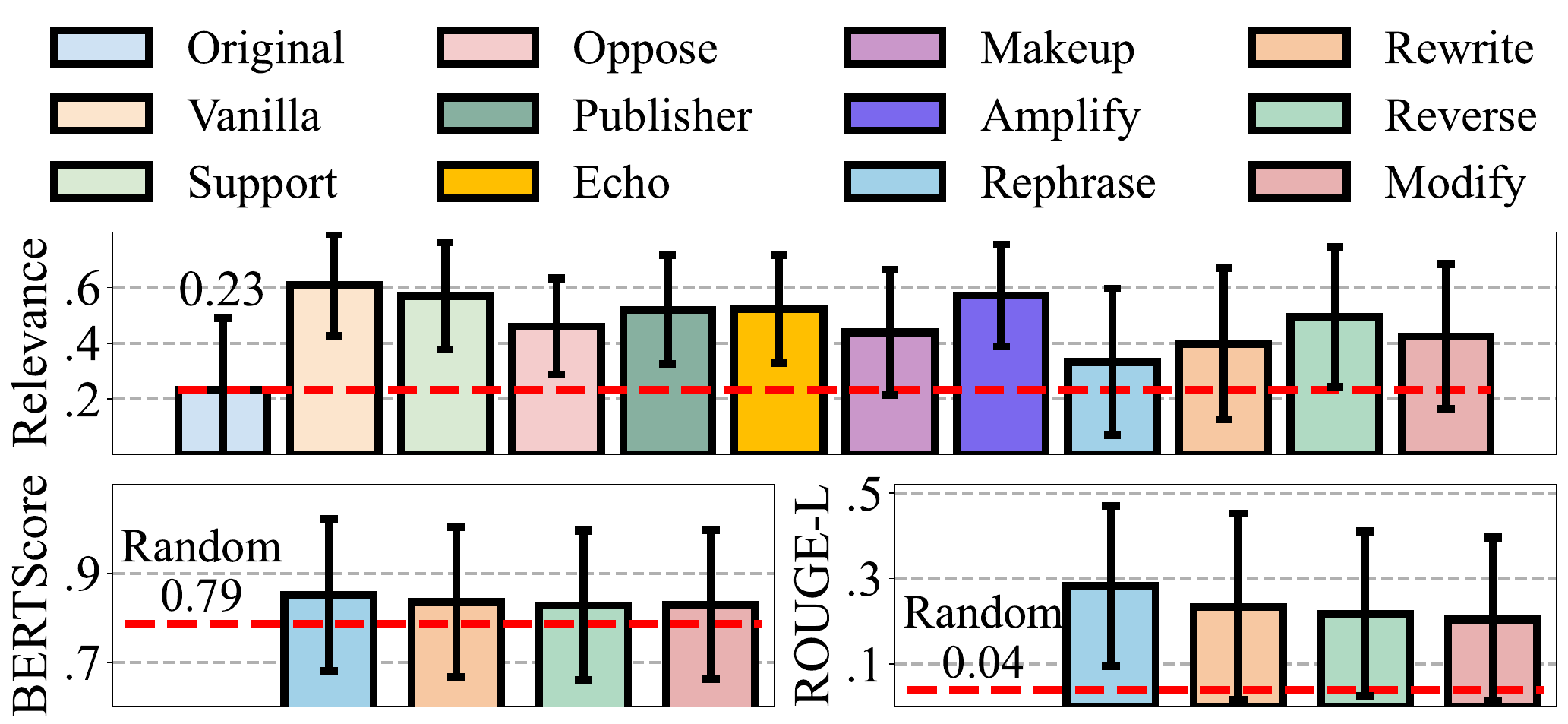}
    \caption{Evaluation of the manipulated evidence. We evaluate the relevance between social text and corresponding evidence and the semantic-level and word-level similarity between original and rephrased evidence. The polluted evidence is of high quality.}
    \label{fig: quality}
\end{figure}

\paragraph{(\MakeUppercase{\romannumeral 1}) The manipulated evidence is of high quality.} We employ SimCSE \citep{gao2021simcse} to evaluate the relevance between social text and corresponding evidence and employ BERTscore \citep{DBLP:conf/iclr/ZhangKWWA20} and ROUGE-L \citep{lin2004rouge} to evaluate the semantic-level and word-level similarity between original and rephrased evidence. Figure \ref{fig: quality} illustrates that the relevance of polluted evidence even exceeds the original. The \textbf{Generate} with an average value of 0.528 is higher than the \textbf{Rephrase} with an average value of 0.412. We speculate that LLMs could follow instructions to generate related evidence, while humans tend to express their opinions unrelated to the original text. Meanwhile, the rephrased evidence is similar to the original in both semantic and word levels, with higher similarities than the randomly selected evidence pairs. We further conduct a human evaluation to check which types of evidence are of high quality. The results show that 12 out of 29 prefer generated evidence to the original, and 14 out of 29 prefer rephrased evidence to the original. We speculate that online social users struggle to distinguish manipulated and original evidence, especially the rephrased type.

\begin{figure}[t]
    \centering
    \includegraphics[width=\linewidth]{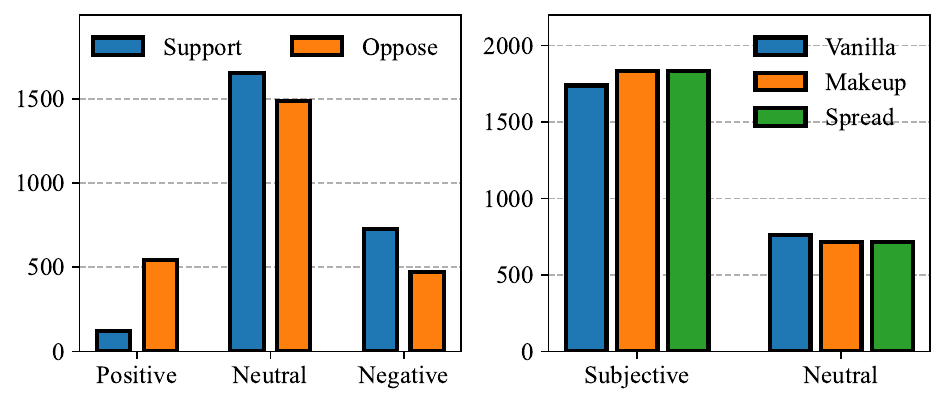}
    \caption{The sentiment and tone distributions of evidence polluted by distinct pollution strategies. LLMs could successfully manipulate evidence to present distinct attributes.}
    \label{fig: differ}
\end{figure}

We further evaluate the differences between the outputs of different strategies. Qualitatively, we provide some cases in Tables \ref{tab: case_1} and \ref{tab: case_2}. It illustrates that LLMs could successfully manipulate the evidence with malicious intent. For example, \textbf{Rephrase} is the most faithful strategy, which does not add fake details. While the other \textbf{Rephrase} strategies will add details that may not be true. Quantitatively, we employ a sentiment and a tone classifier to evaluate the sentiment and style of manipulated evidence. Specifically, we leverage a three-class sentiment classifier (\emph{positive}, \emph{negative}, and \emph{neutral}) from \href{https://huggingface.co/tabularisai/robust-sentiment-analysis}{this link} and a binary tone classifier (\emph{subjective} and \emph{neutral}) from \href{https://huggingface.co/cffl/bert-base-styleclassification-subjective-neutral}{this link}. Figure \ref{fig: differ} presents the results. From a sentiment perspective, we explore the manipulated evidence under the \textbf{Stance} strategy (\textbf{Support} and \textbf{Oppose}) in the \textbf{Pheme} dataset. It illustrates that even if the prompts are similar (with only one word difference), LLMs can generate different comments. From a tone perspective, we explore the manipulated evidence under the \textbf{Vanilla}, \textbf{Makeup}, and \textbf{Spread} in the \textbf{Pheme} dataset. It illustrates that to make information spread fast, LLMs would generate more subjective comments.

\begin{figure}[t]
    \centering
    \includegraphics[width=\linewidth]{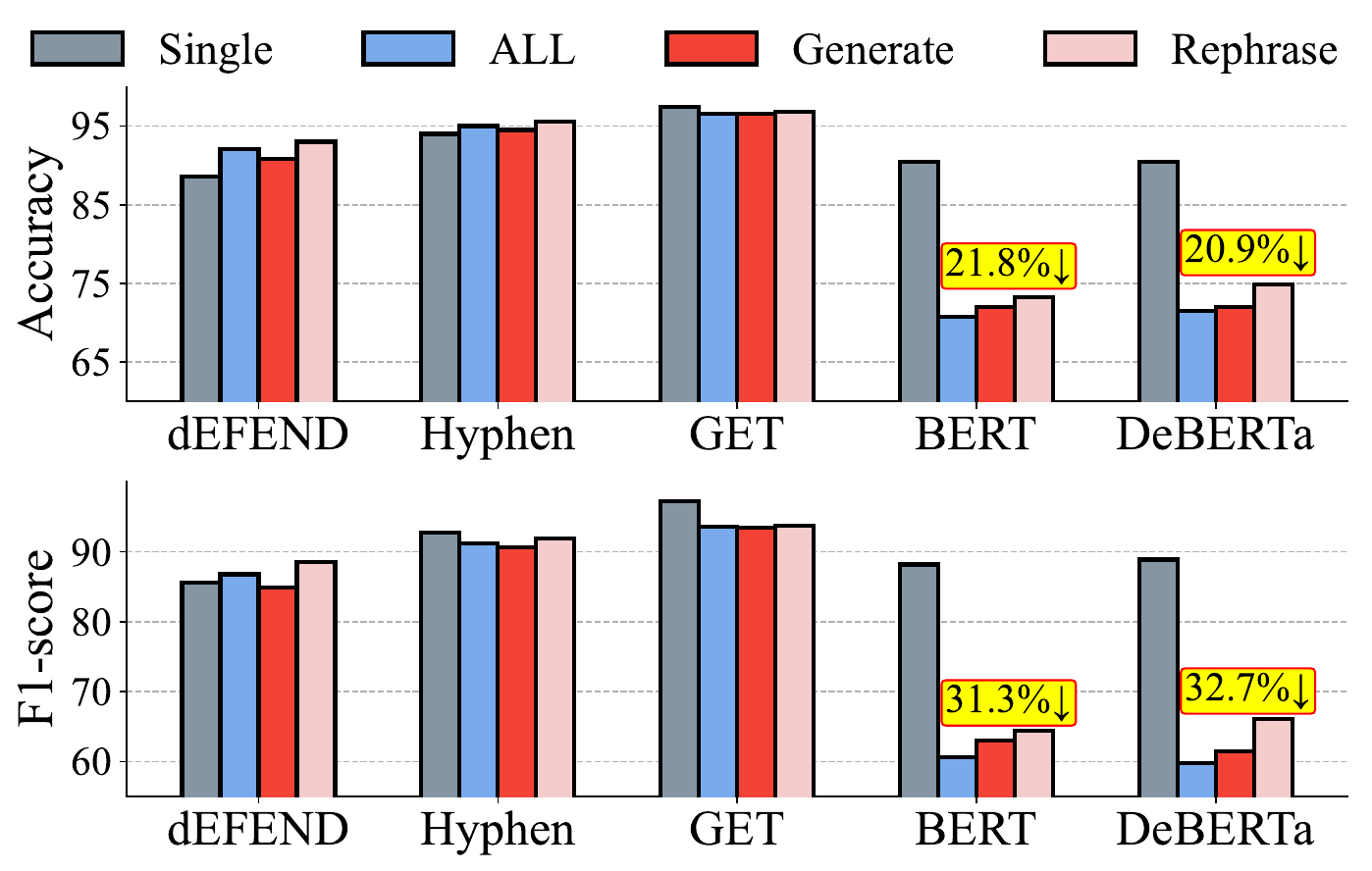}
    \caption{Performance of detectors when the pollution strategies collaborate. For short, ``Single'' denotes the best pollution strategies for a specific detector, ``ALL'' denotes the ensemble of all LLM-based strategies, and ``Generate'' and ``Rephrase'' denote the ensemble of corresponding strategies. The ensemble of evidence pollution would amplify the negative impact.}
    \label{fig: pollution_ensemble}
\end{figure}

\paragraph{(\MakeUppercase{\romannumeral 2}) Evidence pollution harms model calibration thus declining prediction trustworthiness.} Robust detectors should provide a prediction and a well-calibrated confidence score to facilitate content moderation. We evaluate how well detectors are calibrated with original and polluted evidence using Expected Calibration Error (ECE) \citep{guo2017calibration}. Figure \ref{fig: calibration} illustrates partial results, and we present more results in Figure \ref{fig: calibration_appendix} in Appendix \ref{app: more_analysis}. It is demonstrated that polluted evidence harms calibration and increases ECE by up to 21.6\%, while encoder-based LMs are the most well-calibrated.

\paragraph{(\MakeUppercase{\romannumeral 3}) The ensemble of evidence pollution would amplify the negative impact.} Figure \ref{fig: pollution_ensemble} illustrates the performance of detectors when the pollution strategies collaborate. Encoder-based LMs are more sensitive to the ensemble, where \textsc{BERT} drops up to 21.8\% and \textsc{DeBERTa} drops up to 20.9\% for accuracy. Other detectors are more robust but also suffer from slight performance drops.

\section{Related Work}
Identifying malicious social text is critical for ensuring online safety. Researchers work on detecting fake news \citep{yue2023metaadapt, mendes2023human, ma2024event}, identifying rumors \citep{kim2023covid,yang-etal-2024-reinforcement}, countering hate speech \citep{singh2024generalizable, tonneau-etal-2024-naijahate, lee2024exploring}, and recognizing sarcasm \citep{min2023just, chen-etal-2024-cofipara}. Intuitive works employ technologies such as augmentation \citep{kim-etal-2024-label, lee2024exploring}, recurrent neural networks \citep{shu2019defend}, and transformer \citep{tian2023dynamic,nguyen2024vihatet5} enhanced with emotion \citep{zhang2021mining}, opinions \citep{zong2024unveiling}, semantics \citep{ahn-etal-2024-sharedcon}, and logical rules \citep{clarke2023rule, chen2023causal} to analyze social text content. To counter disguised content, evidence-enhanced models are proposed, utilizing external knowledge such as similar content \citep{sheng2022zoom, qi2023two}, comments \citep{yu2023fine, yang2023wsdms}, user \citep{shu2018understanding,dou2021user}, and multiply modalities \citep{cao2020exploring, tiwari2023predict} and then employing networks like graph neural networks \citep{ghosh2023cosyn,jing2023multi} to fuse them. 

Aside from remarkable abilities to standard NLP tasks, LLMs show great potential to conduct content moderation, such as countering social bot detection \citep{feng-etal-2024-bot}, misinformation \citep{russo2023countering, yue2024evidence,ma2024simulated, liu2024teller, su2024adapting}, hate speech \citep{nguyen2023towards,yadav-etal-2024-tox,zheng2024hatemoderate}. However, LLMs' misuse introduces risks of malicious text generation \citep{pelrine2023towards,huang2023faking, DBLP:conf/iclr/ChenS24, wu2024fake}. Existing research explores the influence of misinformation \citep{pan2023risk, goldstein2023generative, xu-etal-2024-earth} and how to detect machine-generated text \citep{mitchell2023detectgpt}. We explored the risks of evidence pollution in malicious social text detection and potential defense strategies, bridging the gap between existing works.

\section{Conclusion}
We explore LLMs' potential evidence pollution risks, which confuse evidence-enhanced malicious social text detectors. We design three types of manipulation strategies including thirteen methods and propose three defense strategies from both the data and model sides. Extensive experiments illustrate that evidence pollution poses a profound threat, which remains challenging to fully mitigate by employing existing defense strategies.

\section*{Acknowledgements}
This work was supported by the National Nature Science Foundation of China (No. 62192781, No. 62272374), the Natural Science Foundation of Shaanxi Province (2024JC-JCQN-62), the National Nature Science Foundation of China (No. 62202367, No. 62250009), the Key Research and Development Project in Shaanxi Province No. 2023GXLH-024, Project of China Knowledge Center for Engineering Science and Technology, and Project of Chinese academy of engineering “The Online and Offline Mixed Educational Service System for ‘The Belt and Road’ Training in MOOC China”, and the K. C. Wong Education Foundation.

\section*{Limitation}
While our proposed pollution strategies and defense strategies are generic, we focus on the comments, which are the most widely used. We believe the extensive experiment results on ten datasets across four malicious social text detection tasks could demonstrate our key contributions.

More recent works might employ the evidence graphs, such as the comments on other comments or user following graphs, to enhance detection performance. This paper focuses on the comments directly on the social text and the textual information instead of graph information. We also believe the extensive experiments of seven strong detectors could demonstrate our key contributions.

We expect to explore the risks of LLMs in manipulating other types of evidence and graph structure, as well as the corresponding defense strategies.

\section*{Ethics Statement}
Identifying malicious social text on social platforms ensures online safety. This paper aims to explore the risks of LLMs in manipulating evidence to compromise evidence-enhanced detectors and develop potential defense strategies to mitigate evidence pollution, while also increasing the risks of dual use. We aim to mitigate such dual use by employing controlled access to our research data, making sure that the data is only employed for research purposes. Meanwhile, our research reveals the vulnerability of existing detectors to evidence pollution. Thus we argue that the decision of the existing detectors should be considered as an initial screen of malicious content, while content moderation decisions should be made with related experts.

We argue that before employing evidence to enhance malicious social text detection, fact-checking is needed to ensure the credibility of the evidence. Meanwhile, to increase the reliability of evidence-enhance detectors, increasing the explainability, such as giving out which evidence leads to the predictions, is critical.

We mainly employ LLMs to rewrite existing evidence or generate fabricated evidence with predetermined malicious intent to compromise detectors. We do not directly employ LLMs to generate malicious content, and we also argue that LLMs should not be employed to generate malicious content, where researchers should make an effort to limit it. Meanwhile, due to the inherent social bias and hallucinations of LLMs, the polluted evidence inevitably contains biased content, such as hate speech or misinformation. We emphasize that the data can only be used for research purposes.

\bibliography{custom}

%\newpage
\appendix

\section{Case Study}
\label{app: case}
We present case studies of each evidence pollution strategy in Tables \ref{tab: case_1} and \ref{tab: case_2}. Note that these cases are all malicious social texts from the public dataset \textbf{Politifact}, and we have concealed personal private information and hate speech as much as possible. We could summarize the characteristics of each evidence pollution strategy as follows:
\begin{itemize}[topsep=4pt, leftmargin=*]
    \item \textbf{Remove} simply removes some related evidence, where the removed evidence might provide useful signals to identify the malicious content. It is straightforward but difficult to implement in practice due to platform rules.
    \item \textbf{Repeat} aims to repeat unified evidence to amplify its influence. It is easily detected by the platforms through the text-matching algorithm.
    \item \textbf{Rephrase} rephrases existing evidence without any additional intents. It is just like a baseline for \textbf{Rephrase Evidence}.
    \item \textbf{Rewrite} rewrites existing evidence intending to make the corresponding social text like a normal one. Thus, LLMs might generate some clarifications in the evidence.
    \item \textbf{Reverse} reverses the stance in existing evidence, thus it might completely replace the content related to the stance.
    \item \textbf{Modify} adds fabricated facts to make the social text human-like. 
    \item \textbf{Vanilla} simply generates related evidence of the corresponding social text. It is just like a baseline for \textbf{Generate Evidence}.
    \item \textbf{Support} generates evidence with the predetermined support stance.
    \item \textbf{Oppose} generates evidence with the predetermined opposing stance.
    \item \textbf{Publisher} simulates the social text publishers to post comments to promote the original social text. For example, LLMs could generate some hashtags.
    \item \textbf{Echo} aims to create echo chambers, where it would post comments with similar semantics. It might be more difficult to be detected by the platforms.
    \item \textbf{Makeup} generates evidence intending to make the corresponding social text like a normal one.
    \item \textbf{Amplify} aims to generate evidence to promote the spread of corresponding social text. Thus LLMs might generate hashtags and employ interrogative sentences.
\end{itemize}

These cases show that the polluted evidence is of high quality, where LLMs could follow the instructions to rewrite or generate highly relevant evidence, confusing existing evidence-enhanced malicious social text detectors.

\section{Metric Set}
\label{app: metrics}
We mainly employ accuracy, macro f1-score, AR\textsubscript{acc} and AR\textsubscript{F1}, and AUC as metrics. We introduce each of the metrics and the reasons to employ them:
\begin{itemize}[topsep=4pt, leftmargin=*]
    \item Accuracy and macro f1-score are widely used metrics for classification tasks. Thus we employ them to evaluate the general performance of detectors. For the accuracy, we employ it in Tables \ref{tab: acc}, \ref{tab: moe_defend}, \ref{tab: moe_get}, \ref{tab: moe_bert}, and \ref{tab: moe_deberta}, and in Figures \ref{fig: updating}, \ref{fig: pollution_ensemble}, \ref{fig: main_strong}, \ref{fig: main_lm}, \ref{fig: main_llm}, \ref{fig: updating_defend}, \ref{fig: updating_hyphen}, \ref{fig: updating_get}, \ref{fig: updating_bert}, and \ref{fig: updating_deberta}. For the macro f1-score, we employ it in Tables \ref{tab: f1}, \ref{tab: moe_defend}, \ref{tab: moe_get}, \ref{tab: moe_bert}, and \ref{tab: moe_deberta}, and Figure \ref{fig: pollution_ensemble}.
    \item AR\textsubscript{acc} and AR\textsubscript{F1} are proposed to evaluate the influence of pollution strategies. Given a specific detector and a pollution strategy, we assume the original performance (accuracy or macro f1-score) is $\{f_i\}_{i=1}^N$, where $N$ is the number of datasets (we employ 10 datasets), and the performance after pollution is $\{\tilde{f}_i\}_{i=1}^N$. The AR is calculated as:
    \begin{align*}
        \mathrm{AR}=\frac{1}{N}\sum_{i=1}^N\frac{\tilde{f}_i}{f_i}.
    \end{align*}
    The lower the value, the more effective the pollution strategy is. Meanwhile, given an AR score, it is convenient to calculate the relative performance drop rate: $1-\mathrm{AR}$. We employ AR in Tables \ref{tab: pollution} and \ref{tab: moe}.
    \item AUC is widely used in machine-generated text detection, thus we employ it to evaluate the performance of machine-generated text detectors, as well as the f1-score. We employ them in Table \ref{tab: machine} and Figure \ref{fig: machine_out}.
\end{itemize}

\section{Baselines}
\label{app: baseline}
We evaluate our proposed evidence pollution and defense strategies on three distinctive types of competitive detectors. The first category is \textbf{existing strong detector}, which presents the most advanced technologies, and we employ:
\begin{itemize}[topsep=4pt, leftmargin=*]
    \item \textsc{dEFEND} \citep{shu2019defend} conducts explainable detection by the attention weights between social text sentences and related evidence. We set the max sentence count of the social text as 8 and the max token count of each sentence as 128. We further set the max evidence count as 10 and the max token count of evidence as 128.
    \item \textsc{Hyphen} \citep{grover2022public} is a discourse-aware hyperbolic spectral co-attention network. It employs a novel Fourier co-attention mechanism to enhance hyperbolic graph representations, obtaining joint representations of social text and evidence. We set the max evidence to count as 10 and the max token count of social text sentence as 128. We further set the max social text sentence count as the 80th percentile for each dataset.
    \item \textsc{GET} \citep{xu2022evidence} models social text and evidence as networks and captures the long-distance semantic dependency among dispersed relevant snippets via neighborhood propagation. For both social text and evidence graphs, we set the max word length as 3840 and set the window size as 5.
\end{itemize}
The second category is \textbf{encoder-based LM}, where we employ encoder-based LMs to encode social text and evidence content and then fuse their representations to conduct classification. Specifically, given a piece of social text $s$ and its corresponding evidence $\{c_i\}_{i=1}^m$, we first employ encoder-based LMs $\mathrm{enc}(\cdot)$ to obtain their representations, \emph{i.e.},
\begin{align*}
    \mathbf{h}_{\textit{text}} &= \mathrm{enc}(s),\\
    \mathbf{h}_{\textit{evid}} &= \sum_{i=1}^m \mathrm{enc}(c_i).
\end{align*}
We then concatenate them to obtain the final representation:
\begin{align*}
    \mathbf{h} = \mathbf{h}_{\textit{text}}\|\mathbf{h}_{\textit{evid}}.
\end{align*}

\begin{table}[t]
    \centering
    \resizebox{\linewidth}{!}{
    \begin{tabular}{c|c|l}
    \toprule[1.5pt]
         Methods&TASK&Prompt\\
    \midrule[1pt]
    \midrule[1pt]
    \multicolumn{3}{l}{Generic input prompt: \textit{Text: }$s$}\\\hline
         \multirow{4}{*}{F3 \textit{VaN}}&Fake News&\textit{Analyze the given text and determine if it is real or fake news.}\\
         &Hate Speech&\textit{Analyze the given text and determine if it is hate speech or not.}\\
         &Rumor&\textit{Analyze the given text and determine if it is a rumor or not a rumor.}\\
         &Sarcasm&\textit{Analyze the given text and determine if it is sarcasm or not.}\\
    \midrule[1pt]
    \midrule[1pt]
    \multicolumn{3}{l}{Generic input prompt: \textit{Text: }$s$ \textit{Comments:} $i$.$c_i$. \textit{Analyze the given text and related comments,} }\\\hline
         \multirow{4}{*}{\textit{w/ evidence}}&Fake News&\textit{and determine if it is real or fake news.}\\
         &Hate Speech&\textit{and determine if it is hate speech or not.}\\
         &Rumor&\textit{and determine if it is a rumor or not a rumor.}\\
         &Sarcasm&\textit{and determine if it is sarcasm or not.}\\
    \midrule[1pt]
    \bottomrule[1.5pt]
    \end{tabular}
    }
    \caption{Prompts of \textbf{LLM-based detectors}, we prompt LLMs using F3 \citep{lucas2023fighting} and with evidence.}
    \label{tab: baseline_prompt}
\end{table}
Finally, given an instance and its label $y$, we compute the probability of $y$ being the correct prediction as $p(y \mid \mathcal{G}) \propto \exp(\mathrm{MLP}(\mathbf{h}))$, where $\mathrm{MLP}(\cdot)$ denotes an MLP layer. We optimize models using the cross-entropy loss and predict the most plausible label as $\arg \max_{y}p(y \mid \mathcal{G})$. In practice, we employ two widely-used encoder-based LMs: (\romannumeral 1) \textsc{BERT} \citep{devlin2019bert} and (\romannumeral 2) \textsc{DeBERTa} \citep{DBLP:conf/iclr/HeGC23}. For LMs without evidence, we directly consider $\mathbf{h}_{\textit{text}}$ as $\mathbf{h}$.

The third category is \textbf{LLM-based detector}, where we prompt LLMs with F3 \citep{lucas2023fighting} and evidence. The detailed prompts are presented in Table \ref{tab: baseline_prompt}. In practice, we employ an open-sourced LLM \textsc{Mistral} and a close-sourced LLM \textsc{ChatGPT}.

\section{Experiment Settings}
\label{app: settings}

\begin{table}[t]
    \centering
    \resizebox{\linewidth}{!}{
    \begin{tabular}{c|c|c|c|c|c}
    \toprule[1.5pt]
         Hyper&\textsc{dEFEND}&\textsc{Hyphen}&\textsc{GET}&\textsc{BERT}&\textsc{DeBERTa}\\
    \midrule[1pt]
         Optimizer&\multicolumn{5}{|c}{Adam (RiemannianAdam for \textsc{Hyphen})}\\
         Metrics&\multicolumn{5}{|c}{Accuracy}\\
         Weight Decay&\multicolumn{5}{|c}{1e-5}\\
         Dropout&\multicolumn{5}{|c}{0.5}\\
         Hidden Dim&\multicolumn{5}{|c}{256}\\\cline{2-6}
         Learning Rate&1e-4&1e-3&1e-3&1e-4&1e-4\\
         Batch Size&32&32&32&16&16\\\hline
         \multicolumn{6}{l}{Only for \textbf{Politifact}, \textbf{Gossipcop}, and \textbf{RumorEval}.}\\\hline
         Batch Size&32&32&32&16&4\\
    \bottomrule[1.5pt]
    \end{tabular}
    }
    \caption{Hyperparameters of baselines required to train.}
    \label{tab: hyper}
\end{table}
\subsection{Baseline Settings}
For each baseline, we conduct ten-fold cross-validation on each dataset to obtain more robust results. We set the hyperparameters the same for each fold. Meanwhile, we run each fold five times and select the checkpoint with the best performance. For each run, we stop training when the performance on the test set does not improve for five epochs. We present the hyperparameters of existing strong detectors and encoder-based LMs in Table \ref{tab: hyper}. For LLM-based Detectors, we set the max new token to count as 50 and set the temperature as zero to obtain fixed predictions.
\subsection{Dataset Settings}
We employ four malicious social text detection tasks including 10 datasets, \emph{i.e.}, (\romannumeral 1) \textbf{fake news detection}: Politicalfact, Gossipcop \citep{shu2020fakenewsnet}, and ANTiVax \citep{hayawi2022anti}; \textbf{hate speech detection}: HASOC \citep{mandl2019overview}; (\romannumeral 3) \textbf{rumor detection}: Pheme \citep{buntain2017automatically}, Twitter15, Twitter16 \citep{ma2018rumor}, and RumorEval \citep{derczynski2017semeval}; (\romannumeral 4) \textbf{sarcasm detection}: Twitter and Reddit \citep{ghosh2020report}. 

\begin{table}[t]
    \centering
    \resizebox{\linewidth}{!}{
    \begin{tabular}{c|c|c|c|c}
    \toprule[1.5pt]
        \textbf{Task}&\textbf{Dataset}&\# Text&\# Malicious&Average \# Evidence \\
    \midrule[1pt]
        \multirow{3}{*}{\textbf{Fake News}}&\textbf{Politifact}&415&270&7.9\\
        &\textbf{Gossipcop}&2,411&1,408&7.6\\
        &\textbf{AnTiVax}&3,797&932&3.6\\
    \midrule[1pt]
        \multirow{1}{*}{\textbf{Hate Speech}}&\textbf{HASOC}&712&298&2.6\\
    \midrule[1pt]
        \multirow{4}{*}{\textbf{Rumor}}&\textbf{Pheme}&6,425&2,402&7.2\\
        &\textbf{Twitter15}&543&276&4.5\\
        &\textbf{Twitter16}&362&163&7.2\\
        &\textbf{RumorEval}$^\star$&446&138&8.1\\
    \midrule[1pt]
        \multirow{2}{*}{\textbf{Sarcasm}}&\textbf{Twitter}&5000&2500&3.6\\
        &\textbf{Reddit}&4400&2200&2.5\\
    \bottomrule[1.5pt]
    \end{tabular}
    }
    \caption{The statistics of the datasets. $\star$ denotes that this dataset contains additional ``not verified'' class.}
    \label{tab: dataset}
\end{table}

For original content and corresponding evidence, we employ the processed data from \textsc{Hyphen} \citep{grover2022public}. We randomly split them into 10 folds to support a ten-fold evaluation. To adapt to each detector and ensure a fair comparison, we randomly down-sample relevant evidence for each social text instance, where each instance contains at most ten pieces of evidence. Table \ref{tab: dataset} presents statistics of the datasets.

\begin{table*}[t]
    \centering
    \resizebox{\textwidth}{!}{
    \begin{tabular}{l|ccc|c|cccc|cc}
    \toprule[1.5pt]
        \multirow{2}{*}{\textbf{Method}}&\multicolumn{3}{c|}{Fake News}&\multicolumn{1}{c|}{Hate Speech}&\multicolumn{4}{c|}{Rumor}&\multicolumn{2}{c}{Sarcasm}\\
        &Politifact&Gossipcop&ANTiVax&HASOC&Pheme&Twitter15&Twitter16&RumorEval&Twitter&Reddit\\
    \midrule[1pt]
        \textsc{dEFEND} \citep{shu2019defend}&$81.4_{\pm 5.1}$&$70.7_{\pm 2.4}$&$90.1_{\pm 1.8}$&$68.4_{\pm 4.2}$&$79.6_{\pm 0.9}$&$84.4_{\pm 4.2}$&$90.6_{\pm 2.8}$&$57.6_{\pm 3.5}$&$75.0_{\pm 1.8}$&$66.2_{\pm 1.3}$\\
        \textsc{Hyphen} \citep{grover2022public}&$88.0_{\pm 6.2}$&$69.1_{\pm 2.5}$&$90.6_{\pm 1.8}$&$67.9_{\pm 7.6}$&$81.0_{\pm 1.3}$&$\underline{90.3}_{\pm 5.3}$&$93.1_{\pm 3.2}$&$63.2_{\pm 5.0}$&$75.5_{\pm 2.0}$&$67.6_{\pm 2.2}$\\
        \textsc{GET} \citep{xu2022evidence}&$93.5_{\pm 4.8}$&$74.3_{\pm 2.3}$&$91.3_{\pm 0.7}$&$66.9_{\pm 5.1}$&$84.8_{\pm 1.5}$&$\textbf{92.2}_{\pm 2.5}$&$\textbf{94.8}_{\pm 3.3}$&$63.7_{\pm 5.2}$&$74.1_{\pm 1.5}$&$65.9_{\pm 2.2}$\\
    \midrule[1pt]
        \textsc{BERT} \citep{devlin2019bert}&$94.0_{\pm 2.9}$&$\underline{76.3}_{\pm 1.8}$&$93.2_{\pm 1.5}$&$\textbf{71.4}_{\pm 4.7}$&$\underline{85.4}_{\pm 1.3}$&$88.5_{\pm 3.8}$&$\underline{93.8}_{\pm 4.4}$&$\textbf{69.0}_{\pm 4.9}$&$\underline{81.0}_{\pm 1.4}$&$70.1_{\pm 1.9}$\\
        \textsc{BERT} w/o comments&$93.1_{\pm 3.7}$&$75.2_{\pm 2.5}$&$92.4_{\pm 1.0}$&$\underline{69.0}_{\pm 5.4}$&$\textbf{86.2}_{\pm 1.8}$&$90.2_{\pm 3.3}$&$\underline{93.8}_{\pm 4.0}$&$\underline{66.1}_{\pm 6.2}$&$79.2_{\pm 1.2}$&$69.7_{\pm 1.8}$\\
        \textsc{DeBERTa} \citep{DBLP:conf/iclr/HeGC23}&$\textbf{96.2}_{\pm 3.5}$&$\textbf{77.3}_{\pm 1.8}$&$\textbf{94.4}_{\pm 1.6}$&$64.7_{\pm 3.1}$&$80.0_{\pm 1.4}$&$83.4_{\pm 4.2}$&$90.0_{\pm 3.9}$&$62.8_{\pm 6.5}$&$\textbf{81.8}_{\pm 1.4}$&$\textbf{73.7}_{\pm 2.1}$\\
        \textsc{DeBERTa} w/o comments&$\underline{96.0}_{\pm 3.4}$&$74.3_{\pm 3.4}$&$\underline{93.9}_{\pm 1.7}$&$62.2_{\pm 5.4}$&$80.9_{\pm 1.0}$&$83.1_{\pm 4.3}$&$91.1_{\pm 4.2}$&$64.9_{\pm 5.8}$&$79.7_{\pm 1.1}$&$\underline{72.7}_{\pm 2.0}$\\
    \midrule[1pt]
        \textsc{Mistral} VaN \citep{lucas2023fighting}&$60.7_{\pm 8.5}$&$33.1_{\pm 2.7}$&$52.8_{\pm 2.2}$&$44.1_{\pm 4.5}$&$47.1_{\pm 1.7}$&$37.7_{\pm 9.4}$&$34.5_{\pm 5.5}$&$30.4_{\pm 10.9}$&$63.0_{\pm 1.7}$&$55.7_{\pm 2.1}$\\
        \textsc{Mistral} w/ comment&$53.2_{\pm 10.1}$&$39.2_{\pm 4.1}$&$36.6_{\pm 2.9}$&$46.0_{\pm 5.1}$&$50.5_{\pm 1.7}$&$36.7_{\pm 6.3}$&$31.1_{\pm 3.1}$&$37.1_{\pm 8.4}$&$59.0_{\pm 2.5}$&$51.6_{\pm 1.6}$\\
        \textsc{ChatGPT} VaN \citep{lucas2023fighting}&$49.3_{\pm 7.5}$&$29.1_{\pm 2.1}$&$45.0_{\pm 2.0}$&$55.6_{\pm 6.0}$&$27.8_{\pm 1.0}$&$39.7_{\pm 5.2}$&$39.3_{\pm 5.6}$&$39.1_{\pm 8.9}$&$40.4_{\pm 2.0}$&$37.1_{\pm 1.9}$\\
        \textsc{ChatGPT} w/ comments&$61.7_{\pm 7.3}$&$29.2_{\pm 2.2}$&$59.4_{\pm 3.9}$&$56.4_{\pm 5.4}$&$31.1_{\pm 1.2}$&$45.6_{\pm 7.7}$&$38.8_{\pm 7.5}$&$23.2_{\pm 6.3}$&$60.2_{\pm 1.8}$&$53.4_{\pm 1.9}$\\
        
    \bottomrule[1.5pt]
    \end{tabular}
    }
    \caption{Macro f1-Score of baselines on ten datasets from four malicious text-related tasks. We conduct ten-fold cross-validation and report the mean and standard deviation to obtain a more robust conclusion. \textbf{Bold} indicates the best performance and \underline{underline} indicates the second best. Evidence could provide valuable signals to enhance detection, however, LLM-based models struggle to detect malicious content.}
    \label{tab: f1}
\end{table*}

\subsection{Evidence Pollution Settings}
We employ \textit{Mistral-7B} \citep{jiang2023mistral} to rephrase and generate polluted evidence. To ensure reproducibility, we set the temperature as zero. For \textbf{Rephrase} strategy, we prompt LLMs to rephrase in three ways, however, we employ the first version in practice because their performance is similar.
\subsection{Defense Strategy Settings}
\paragraph{Machine-Generated Text Detection} To construct datasets for evaluating machine-generated text detectors, we sample 200 pieces of evidence from each dataset on each pollution strategy and original evidence, resulting in 2,000 sentences for each set. We then consider the polluted evidence as machine-generated data and the original evidence as human-written data and mix them, obtaining 11 datasets where each dataset contains 4,000 sentences, named by the pollution strategy, such as \textbf{Rephrase} and \textbf{Support}. We finally split each dataset into the training set, valuation set, and test set by 2:1:1. For \emph{metric-based} methods not required to train, we evaluate it on the test set. We employ roc auc and f1-score as metrics. For DeBERTa-v3, we set batch size as 24, learning rate as 1e-4, optimizer as Adam, weight decay as 1e-5, and hidden dim as 512. For FastGPT, we employ the official implementation\footnote{\url{https://github.com/baoguangsheng/fast-detect-gpt}} to obtain the prediction results.

For out-of-domain evaluation of DeBERTa, we keep the parameters the same and directly evaluate DeBERTa trained on a specific dataset on another.
\paragraph{Mixture of Expert} We set $k$ as $m$, namely, if a specific social text contains $m$ pieces of evidence, then, we consider each piece of evidence as a group, obtaining $m$ groups. Formally, given a detector $f$ and its fixed parameters $\theta$, social text $s$, and its corresponding evidence $\{c_i\}_{i=1}^m$, we could obtain $m$ predictions as:
\begin{align*}
    y_i=\mathop{\arg\max}\limits_{y}p(y\mid s,\{c_i\},f,\theta).
\end{align*}
We  then obtain the final prediction as:
\begin{align*}
    y = \mathop{\arg\max}\limits_{y_j}(\sum_{i=1}^k\mathbf{I}(y_i=y_j)).
\end{align*}

We evaluate this strategy on existing \textbf{strong detectors} and \textbf{encoder-based LMs} except \textsc{Hyphen}. \textsc{Hyphen} extracts the reference relations from multiple pieces of evidence, thus unsuitable for this strategy and would cost huge computation resources. Meanwhile, this strategy is unsuitable for \textbf{LLM-based detectors}, where it would cost huge input tokens. Given $m$ pieces of evidence, the consumed tokens would be increased by $m$ times.
\paragraph{Parameter Updating}
We employ 1\% to 10\% data from the training set to update the model parameters for each dataset, where we set the learning rate as 1e-4, batch size as 5, weight decay as 1e-5, and optimizer as Adam. To simulate the realistic situation that required a quick response, we just re-train the model using the training data only once.

\subsection{Analysis Settings}
\paragraph{Metric-based Evaluation of Polluted Evidence} We first randomly sample 100 instances from each dataset to obtain a generic evaluation. We then calculate the relevant score between social text and corresponding evidence and calculate the BERTScore and ROUGE-L between rephrased and original evidence. For the ``Random'' category, we shuffle the initial polluted-original evidence pairs and consider it as a baseline.

For the relevant scores, we employ the hugging face implementation\footnote{\url{https://huggingface.co/princeton-nlp/sup-simcse-bert-base-uncased}}. For BERTScore, we employ its official implementation\footnote{\url{https://github.com/Tiiiger/bert_score}} and set rescale with baseline as False, and for ROUGE-L, we employ the python packet\footnote{\url{https://pypi.org/project/rouge-score/}}.
\paragraph{Human Evaluation of Polluted Evidence}
We recruit 99 annotators familiar with social networking platforms to judge which comment is of higher quality for a certain social text. For each annotator, we sample 15 generate-original evidence pairs, 15 rephrase-original evidence pairs, 15 generate-rephrase evidence pairs, and 5 randomly shuffled pairs as benchmark questions where the comment with higher quality is clear. We first give each annotator a brief guideline:

\emph{Thank you for attending our human evaluation. Social media users would comment on a post to express their opinions. You are asked to check which comment is of higher quality for a certain post (comment 1 or 2). Please consider factors such as relevance to the post, tone, suitability for the social platform (for the use of hashtags), etc. Please do not consider the length and grammatical errors of the comment. If you think two comments are of equal quality, please subjectively choose the one you like.}

After that, if an annotator correctly identifies 3 out of 5 benchmark questions, we accept his annotations, obtaining 29 annotations.

\paragraph{Calibration Settings}
We consider the max value of the logits after the softmax operator as the confidence scores. For example, if the output is [0.8, 0.2], then the confidence score is 0.8, and if the output is [0.25, 0.75], then the confidence score is 0.75. Figure \ref{fig: calibration} presents the model calibration when the evidence pollution strategies are mixed, while Figure \ref{fig: calibration_appendix} presents the calibration of each pollution strategy.
\paragraph{Pollution Ensemble Settings}
We directly employ majority voting to obtain the ensemble predictions by multiple pollution strategies.

\section{More Results of Evidence Pollution}
\label{app: more_pollution}
Table \ref{tab: f1} presents the macro f1-score of baselines, where it shows a similar trend as accuracy.

\begin{figure*}[t]
    \centering
    \includegraphics[width=0.8\linewidth]{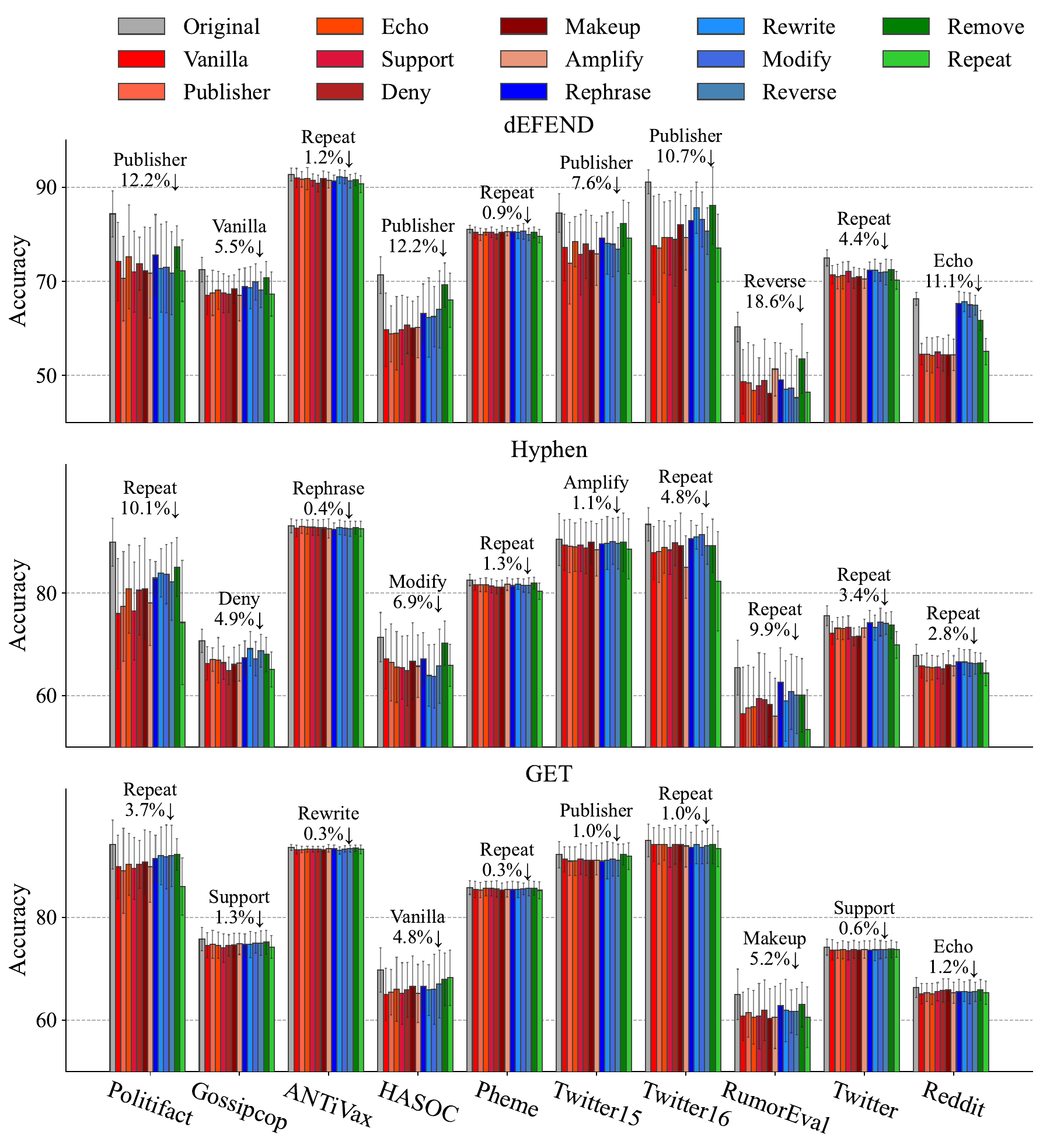}
    \caption{Performance of \textbf{existing strong detectors} on different datasets under different pollution strategies. We illustrate the most effective pollution strategy on each dataset for each model.}
    \label{fig: main_strong}
\end{figure*}

\begin{figure*}[t]
    \centering
    \includegraphics[width=0.8\linewidth]{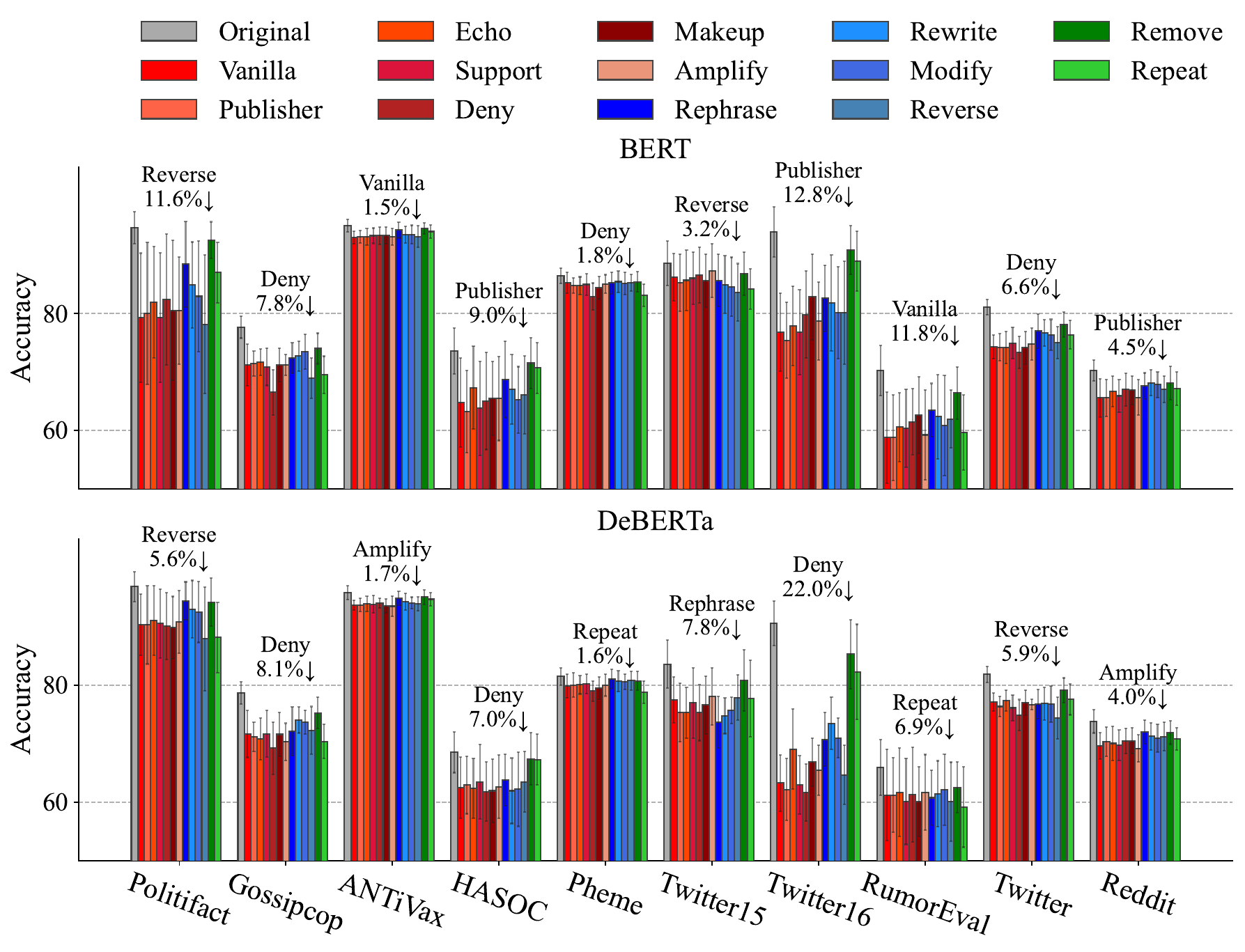}
    \caption{Performance of \textbf{encoder-based LMs} on different datasets under different pollution strategies. We illustrate the most effective pollution strategy on each dataset for each model.}
    \label{fig: main_lm}
\end{figure*}

\begin{figure*}[t]
    \centering
    \includegraphics[width=0.8\linewidth]{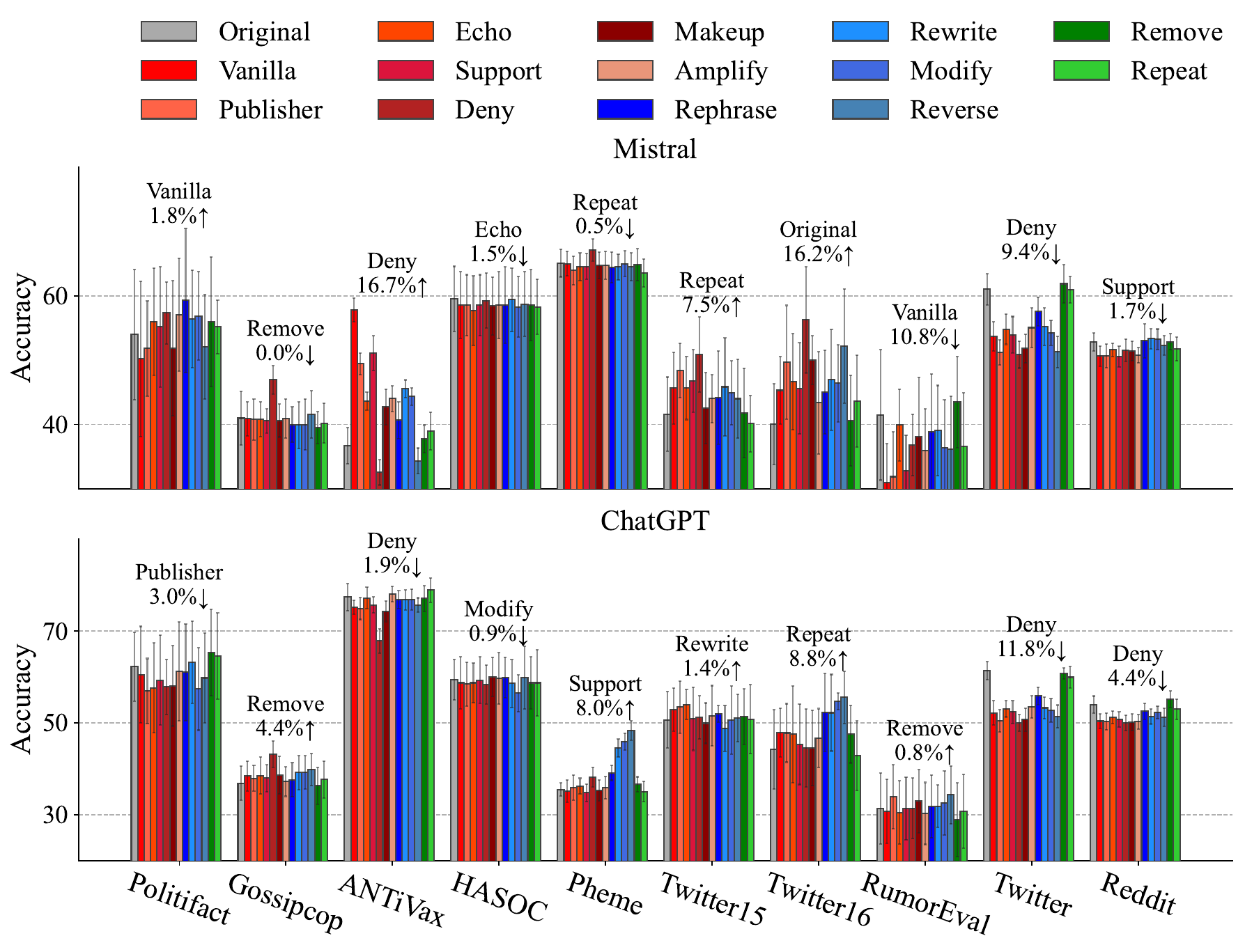}
    \caption{Performance of \textbf{LLM-based detectors} on different datasets under different pollution strategies. We illustrate the most effective pollution strategy on each dataset for each model.}
    \label{fig: main_llm}
\end{figure*}

Meanwhile, we present the whole accuracy of the seven baselines on ten datasets under each pollution strategy in Figures \ref{fig: main_strong}, \ref{fig: main_lm}, and \ref{fig: main_llm}. We only present accuracy because macro-f1 shows similar trends as accuracy shown in Tables \ref{tab: acc} and  \ref{tab: f1}. The additional results strengthen that evidence pollution significantly compromised evidence-enhanced malicious social text detection performance.

\begin{table*}[t]
    \centering
    \resizebox{\linewidth}{!}{
\begin{tabular}{c|cccccc|cc|cccccccc|cccc}
    \toprule[1.5pt]
    \multirow{3}{*}{\textbf{Pollution}}&\multicolumn{6}{c|}{\textbf{Fake News}}&\multicolumn{2}{c|}{\textbf{Hate Speech}}&\multicolumn{8}{c|}{\textbf{Rumor}}&\multicolumn{4}{c}{\textbf{Sarcasm}}\\\cline{2-21}
    &\multicolumn{2}{c}{\textbf{Politifact}}&\multicolumn{2}{c}{\textbf{Gossipcop}}&\multicolumn{2}{c|}{\textbf{ANTiVax}}&\multicolumn{2}{c|}{\textbf{HASOC}}&\multicolumn{2}{c}{\textbf{Pheme}}&\multicolumn{2}{c}{\textbf{Twitter15}}&\multicolumn{2}{c}{\textbf{Twitter16}}&\multicolumn{2}{c|}{\textbf{RumorEval}}&\multicolumn{2}{c}{\textbf{Twitter}}&\multicolumn{2}{c}{\textbf{Reddit}}\\\cline{2-21}
    &Acc&F1&Acc&F1&Acc&F1&Acc&F1&Acc&F1&Acc&F1&Acc&F1&Acc&F1&Acc&F1&Acc&F1\\
    \midrule[1pt]
    \midrule[1pt]
\multirow{3}{*}{\textbf{Vanilla}}&$74.2_{\pm 8.3}$&$70.8_{\pm 7.8}$&\cellcolor{gray!30}$67.0_{\pm 4.1}$&$63.4_{\pm 5.2}$&$92.0_{\pm 2.0}$&$89.1_{\pm 2.7}$&\cellcolor{gray!30}$59.7_{\pm 7.8}$&\cellcolor{gray!30}$50.0_{\pm 10.2}$&$80.4_{\pm 1.2}$&$78.9_{\pm 1.4}$&$77.2_{\pm 7.1}$&$77.0_{\pm 7.2}$&$77.6_{\pm 10.5}$&$76.5_{\pm 11.2}$&$48.6_{\pm 6.8}$&$45.8_{\pm 7.4}$&$71.4_{\pm 1.9}$&$71.4_{\pm 1.9}$&\cellcolor{gray!30}$54.5_{\pm 2.3}$&\cellcolor{gray!30}$47.2_{\pm 6.9}$\\
&$71.5_{\pm 8.2}$&$68.3_{\pm 7.3}$&\cellcolor{gray!30}$67.4_{\pm 4.7}$&$61.5_{\pm 7.0}$&$91.5_{\pm 1.7}$&$88.5_{\pm 2.4}$&\cellcolor{gray!30}$61.9_{\pm 6.5}$&\cellcolor{gray!30}$54.5_{\pm 8.6}$&$80.0_{\pm 1.4}$&$78.6_{\pm 1.5}$&$76.9_{\pm 9.5}$&$76.5_{\pm 9.7}$&$76.5_{\pm 9.7}$&$75.9_{\pm 9.5}$&$46.1_{\pm 10.2}$&$37.2_{\pm 11.5}$&$71.3_{\pm 2.5}$&$71.2_{\pm 2.5}$&\cellcolor{gray!30}$59.9_{\pm 2.2}$&\cellcolor{gray!30}$57.7_{\pm 4.2}$\\
\end{tabular}
    }
    \caption{The \textbf{Mixture of Experts} strategy performance on \textsc{dEFEND}. We \colorbox{gray!30}{highlight} the improved parts.}
    \label{tab: moe_defend}
\end{table*}

\begin{table*}[t]
    \centering
    \resizebox{\linewidth}{!}{
    \begin{tabular}{c|cccccc|cc|cccccccc|cccc}
    \toprule[1.5pt]
    \multirow{3}{*}{\textbf{Pollution}}&\multicolumn{6}{c|}{\textbf{Fake News}}&\multicolumn{2}{c|}{\textbf{Hate Speech}}&\multicolumn{8}{c|}{\textbf{Rumor}}&\multicolumn{4}{c}{\textbf{Sarcasm}}\\\cline{2-21}
    &\multicolumn{2}{c}{\textbf{Politifact}}&\multicolumn{2}{c}{\textbf{Gossipcop}}&\multicolumn{2}{c|}{\textbf{ANTiVax}}&\multicolumn{2}{c|}{\textbf{HASOC}}&\multicolumn{2}{c}{\textbf{Pheme}}&\multicolumn{2}{c}{\textbf{Twitter15}}&\multicolumn{2}{c}{\textbf{Twitter16}}&\multicolumn{2}{c|}{\textbf{RumorEval}}&\multicolumn{2}{c}{\textbf{Twitter}}&\multicolumn{2}{c}{\textbf{Reddit}}\\\cline{2-21}
    &Acc&F1&Acc&F1&Acc&F1&Acc&F1&Acc&F1&Acc&F1&Acc&F1&Acc&F1&Acc&F1&Acc&F1\\
    \midrule[1pt]
    \midrule[1pt]
\multirow{3}{*}{\textbf{Vanilla}}&$89.8_{\pm 6.2}$&$88.5_{\pm 6.8}$&$74.6_{\pm 2.4}$&$72.9_{\pm 3.2}$&\cellcolor{gray!30}$93.2_{\pm 0.8}$&\cellcolor{gray!30}$90.7_{\pm 0.9}$&$65.0_{\pm 5.1}$&$61.6_{\pm 4.8}$&$85.5_{\pm 1.6}$&\cellcolor{gray!30}$84.5_{\pm 1.7}$&\cellcolor{gray!30}$91.3_{\pm 2.5}$&$91.3_{\pm 2.4}$&\cellcolor{gray!30}$94.2_{\pm 3.3}$&\cellcolor{gray!30}$93.9_{\pm 3.6}$&\cellcolor{gray!30}$60.8_{\pm 4.7}$&\cellcolor{gray!30}$59.6_{\pm 4.8}$&$73.7_{\pm 2.0}$&$73.6_{\pm 2.0}$&\cellcolor{gray!30}$65.1_{\pm 2.0}$&\cellcolor{gray!30}$64.8_{\pm 2.2}$\\
&$87.4_{\pm 7.4}$&$85.9_{\pm 7.5}$&$74.5_{\pm 2.2}$&$72.5_{\pm 3.2}$&\cellcolor{gray!30}$93.2_{\pm 0.8}$&\cellcolor{gray!30}$90.8_{\pm 0.9}$&$64.6_{\pm 5.3}$&$61.1_{\pm 5.2}$&$85.4_{\pm 1.6}$&\cellcolor{gray!30}$84.5_{\pm 1.7}$&\cellcolor{gray!30}$91.3_{\pm 3.0}$&$91.3_{\pm 3.0}$&\cellcolor{gray!30}$94.2_{\pm 2.7}$&\cellcolor{gray!30}$93.9_{\pm 3.0}$&\cellcolor{gray!30}$61.4_{\pm 4.9}$&\cellcolor{gray!30}$60.4_{\pm 5.1}$&$73.5_{\pm 1.8}$&$73.4_{\pm 1.8}$&\cellcolor{gray!30}$65.7_{\pm 2.3}$&\cellcolor{gray!30}$65.2_{\pm 2.7}$\\
\end{tabular}
    }
    \caption{The \textbf{Mixture of Experts} strategy performance on \textsc{GET}. We \colorbox{gray!30}{highlight} the improved parts.}
    \label{tab: moe_get}
\end{table*}

\begin{table*}[t]
    \centering
    \resizebox{\linewidth}{!}{
    \begin{tabular}{c|cccccc|cc|cccccccc|cccc}
    \toprule[1.5pt]
    \multirow{3}{*}{\textbf{Pollution}}&\multicolumn{6}{c|}{\textbf{Fake News}}&\multicolumn{2}{c|}{\textbf{Hate Speech}}&\multicolumn{8}{c|}{\textbf{Rumor}}&\multicolumn{4}{c}{\textbf{Sarcasm}}\\\cline{2-21}
    &\multicolumn{2}{c}{\textbf{Politifact}}&\multicolumn{2}{c}{\textbf{Gossipcop}}&\multicolumn{2}{c|}{\textbf{ANTiVax}}&\multicolumn{2}{c|}{\textbf{HASOC}}&\multicolumn{2}{c}{\textbf{Pheme}}&\multicolumn{2}{c}{\textbf{Twitter15}}&\multicolumn{2}{c}{\textbf{Twitter16}}&\multicolumn{2}{c|}{\textbf{RumorEval}}&\multicolumn{2}{c}{\textbf{Twitter}}&\multicolumn{2}{c}{\textbf{Reddit}}\\\cline{2-21}
    &Acc&F1&Acc&F1&Acc&F1&Acc&F1&Acc&F1&Acc&F1&Acc&F1&Acc&F1&Acc&F1&Acc&F1\\
    \midrule[1pt]
    \midrule[1pt]
\multirow{3}{*}{\textbf{Vanilla}}&$79.3_{\pm 11.0}$&$78.2_{\pm 10.4}$&$71.2_{\pm 3.6}$&$70.4_{\pm 3.6}$&\cellcolor{gray!30}$93.0_{\pm 1.1}$&\cellcolor{gray!30}$90.1_{\pm 2.1}$&$64.8_{\pm 7.6}$&$56.8_{\pm 7.9}$&$85.2_{\pm 1.8}$&$84.0_{\pm 1.8}$&\cellcolor{gray!30}$86.2_{\pm 4.1}$&\cellcolor{gray!30}$86.1_{\pm 4.1}$&$76.8_{\pm 6.7}$&$76.1_{\pm 6.9}$&\cellcolor{gray!30}$58.8_{\pm 7.8}$&\cellcolor{gray!30}$54.9_{\pm 8.2}$&$74.3_{\pm 2.0}$&$74.2_{\pm 2.0}$&\cellcolor{gray!30}$65.6_{\pm 3.3}$&\cellcolor{gray!30}$65.2_{\pm 3.5}$\\
&$78.5_{\pm 9.9}$&$77.3_{\pm 9.4}$&$71.0_{\pm 2.7}$&$70.3_{\pm 2.7}$&\cellcolor{gray!30}$93.0_{\pm 1.2}$&\cellcolor{gray!30}$90.1_{\pm 2.1}$&$64.6_{\pm 8.4}$&$56.5_{\pm 8.9}$&$85.1_{\pm 1.7}$&$83.9_{\pm 1.7}$&\cellcolor{gray!30}$86.6_{\pm 4.8}$&\cellcolor{gray!30}$86.5_{\pm 4.9}$&$76.5_{\pm 7.9}$&$75.8_{\pm 8.1}$&\cellcolor{gray!30}$59.2_{\pm 7.4}$&\cellcolor{gray!30}$55.5_{\pm 7.6}$&$74.0_{\pm 1.9}$&$74.0_{\pm 1.9}$&\cellcolor{gray!30}$65.7_{\pm 3.4}$&\cellcolor{gray!30}$65.2_{\pm 3.7}$\\
\end{tabular}
    }
    \caption{The \textbf{Mixture of Experts} strategy performance on \textsc{BERT}. We \colorbox{gray!30}{highlight} the improved parts.}
    \label{tab: moe_bert}
\end{table*}

\begin{table*}[t]
    \centering
    \resizebox{\linewidth}{!}{
    \begin{tabular}{c|cccccc|cc|cccccccc|cccc}
    \toprule[1.5pt]
    \multirow{3}{*}{\textbf{Pollution}}&\multicolumn{6}{c|}{\textbf{Fake News}}&\multicolumn{2}{c|}{\textbf{Hate Speech}}&\multicolumn{8}{c|}{\textbf{Rumor}}&\multicolumn{4}{c}{\textbf{Sarcasm}}\\\cline{2-21}
    &\multicolumn{2}{c}{\textbf{Politifact}}&\multicolumn{2}{c}{\textbf{Gossipcop}}&\multicolumn{2}{c|}{\textbf{ANTiVax}}&\multicolumn{2}{c|}{\textbf{HASOC}}&\multicolumn{2}{c}{\textbf{Pheme}}&\multicolumn{2}{c}{\textbf{Twitter15}}&\multicolumn{2}{c}{\textbf{Twitter16}}&\multicolumn{2}{c|}{\textbf{RumorEval}}&\multicolumn{2}{c}{\textbf{Twitter}}&\multicolumn{2}{c}{\textbf{Reddit}}\\\cline{2-21}
    &Acc&F1&Acc&F1&Acc&F1&Acc&F1&Acc&F1&Acc&F1&Acc&F1&Acc&F1&Acc&F1&Acc&F1\\
    \midrule[1pt]
    \midrule[1pt]
\multirow{3}{*}{\textbf{Vanilla}}&$90.4_{\pm 5.2}$&$89.4_{\pm 6.1}$&\cellcolor{gray!30}$71.7_{\pm 4.0}$&$69.9_{\pm 3.7}$&$93.7_{\pm 0.9}$&$91.1_{\pm 1.6}$&$62.5_{\pm 5.2}$&$55.2_{\pm 4.4}$&$79.8_{\pm 2.0}$&\cellcolor{gray!30}$78.3_{\pm 2.2}$&$77.5_{\pm 3.9}$&$76.8_{\pm 4.7}$&$63.3_{\pm 4.8}$&$59.9_{\pm 5.6}$&\cellcolor{gray!30}$61.2_{\pm 7.8}$&\cellcolor{gray!30}$57.0_{\pm 9.4}$&\cellcolor{gray!30}$77.1_{\pm 1.5}$&\cellcolor{gray!30}$77.0_{\pm 1.5}$&$69.6_{\pm 2.3}$&$69.2_{\pm 2.3}$\\
&$89.2_{\pm 4.6}$&$88.2_{\pm 5.4}$&\cellcolor{gray!30}$71.8_{\pm 3.9}$&$69.7_{\pm 3.8}$&$93.6_{\pm 1.0}$&$90.9_{\pm 1.8}$&$61.8_{\pm 5.2}$&$54.4_{\pm 3.7}$&$79.8_{\pm 2.0}$&\cellcolor{gray!30}$78.3_{\pm 2.1}$&$77.3_{\pm 3.8}$&$76.6_{\pm 4.5}$&$63.0_{\pm 5.5}$&$59.4_{\pm 6.5}$&\cellcolor{gray!30}$62.1_{\pm 8.8}$&\cellcolor{gray!30}$58.1_{\pm 9.8}$&\cellcolor{gray!30}$77.4_{\pm 1.7}$&\cellcolor{gray!30}$77.4_{\pm 1.7}$&$69.4_{\pm 2.6}$&$69.0_{\pm 2.8}$\\
\end{tabular}
    }
    \caption{The \textbf{Mixture of Experts} strategy performance on \textsc{DeBERTa}. We \colorbox{gray!30}{highlight} the improved parts.}
    \label{tab: moe_deberta}
\end{table*}

\section{More Results of Defense Strategies}
\label{app: more_defense}
\subsection{Mixture of Experts}
Tables \ref{tab: moe_defend}, \ref{tab: moe_get}, \ref{tab: moe_bert}, and \ref{tab: moe_deberta} present the performance of \textbf{mixture of experts} of each baseline on different datasets under different pollution strategies. We highlight the values where the strategy imitates the negative impact. The results show that this defense strategy could improve the detection performance on some datasets under some strategies. However, in some cases, this strategy might harm the performance. It strength that although the mixture of experts could improve the performance, it would introduce some noise, declining the performance.
\subsection{Paramter Updating}
\begin{figure*}
    \centering
    \includegraphics[width=0.9\linewidth]{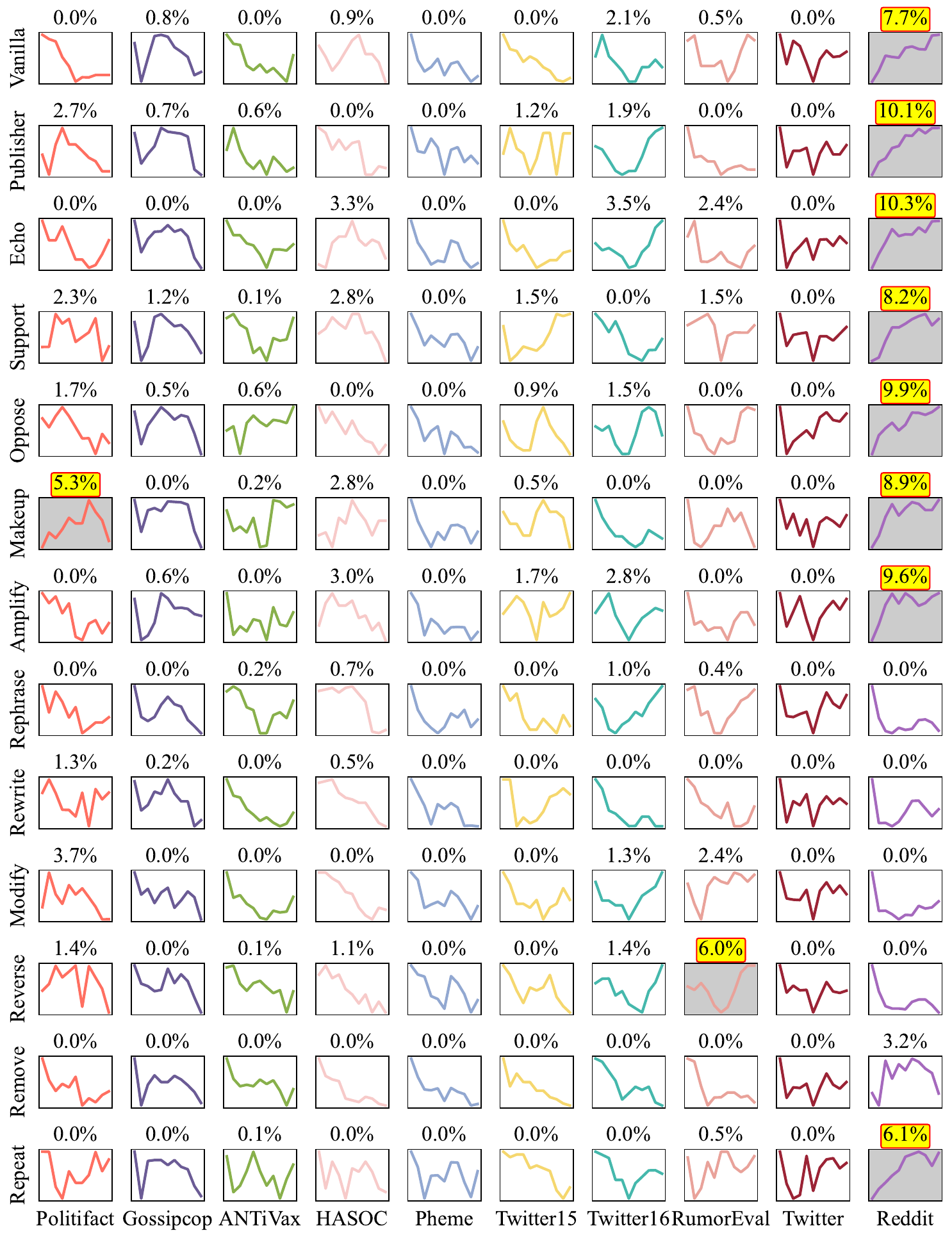}
    \caption{The performance trend of \textbf{Parameter Updating} strategy with re-training data increasing. We present \textsc{dEFEND} on different datasets under different pollution strategies. We present the max improvement of each situation and \colorbox{yellow}{highlight} the top-ten improvement. It strengthens that \textbf{Parameter updating} is the most effective defense strategy, however, the need for annotated data and the unknown when the training ends limit its practical application.}
    \label{fig: updating_defend}
\end{figure*}

\begin{figure*}
    \centering
    \includegraphics[width=0.9\linewidth]{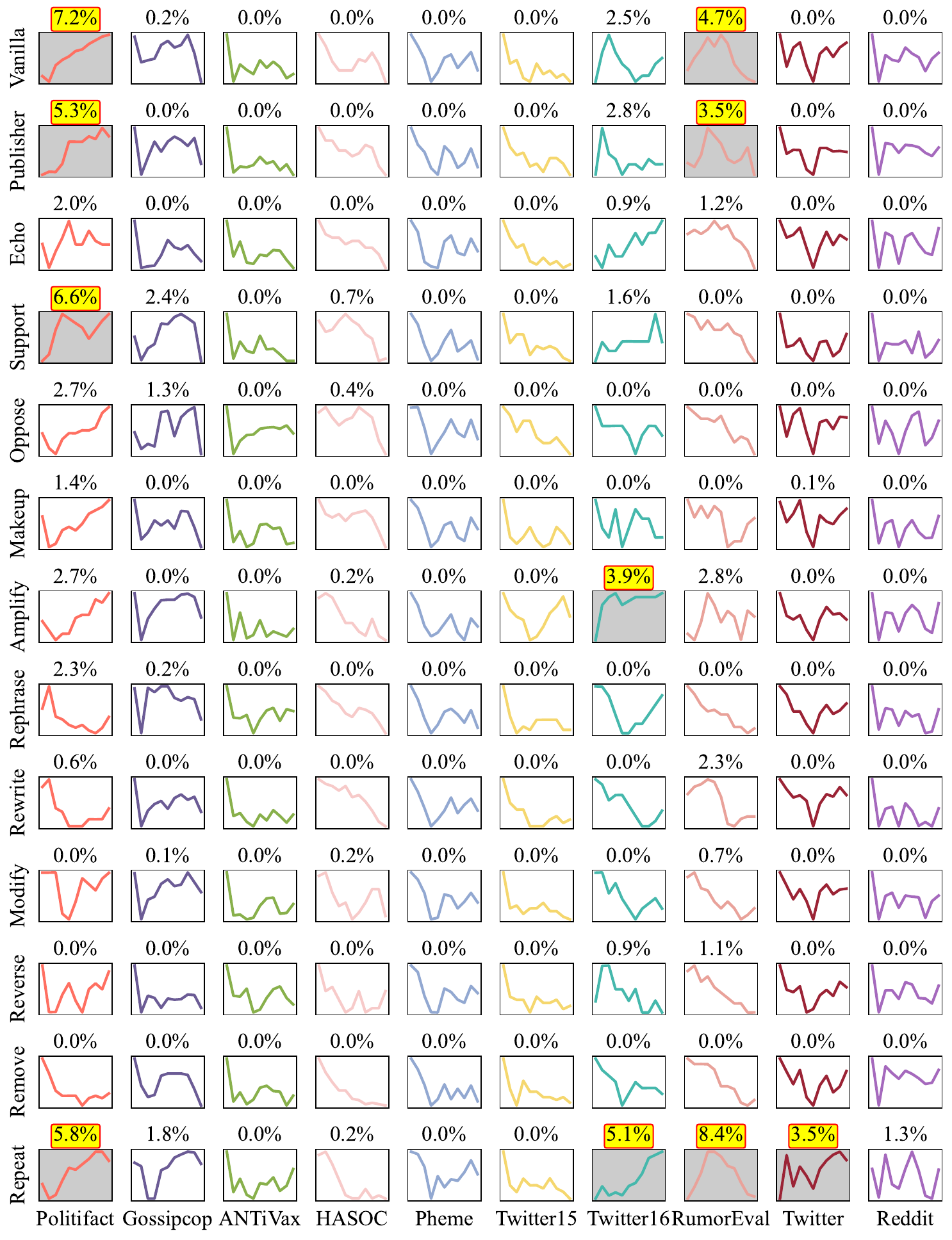}
    \caption{The performance trend of \textbf{Parameter Updating} strategy with re-training data increasing. We present \textsc{Hyphen} on different datasets under different pollution strategies. We present the max improvement of each situation and \colorbox{yellow}{highlight} the top-ten improvement. It strengthens that \textbf{Parameter updating} is the most effective defense strategy, however, the need for annotated data and the unknown when the training ends limit its practical application.}
    \label{fig: updating_hyphen}
\end{figure*}

\begin{figure*}
    \centering
    \includegraphics[width=0.9\linewidth]{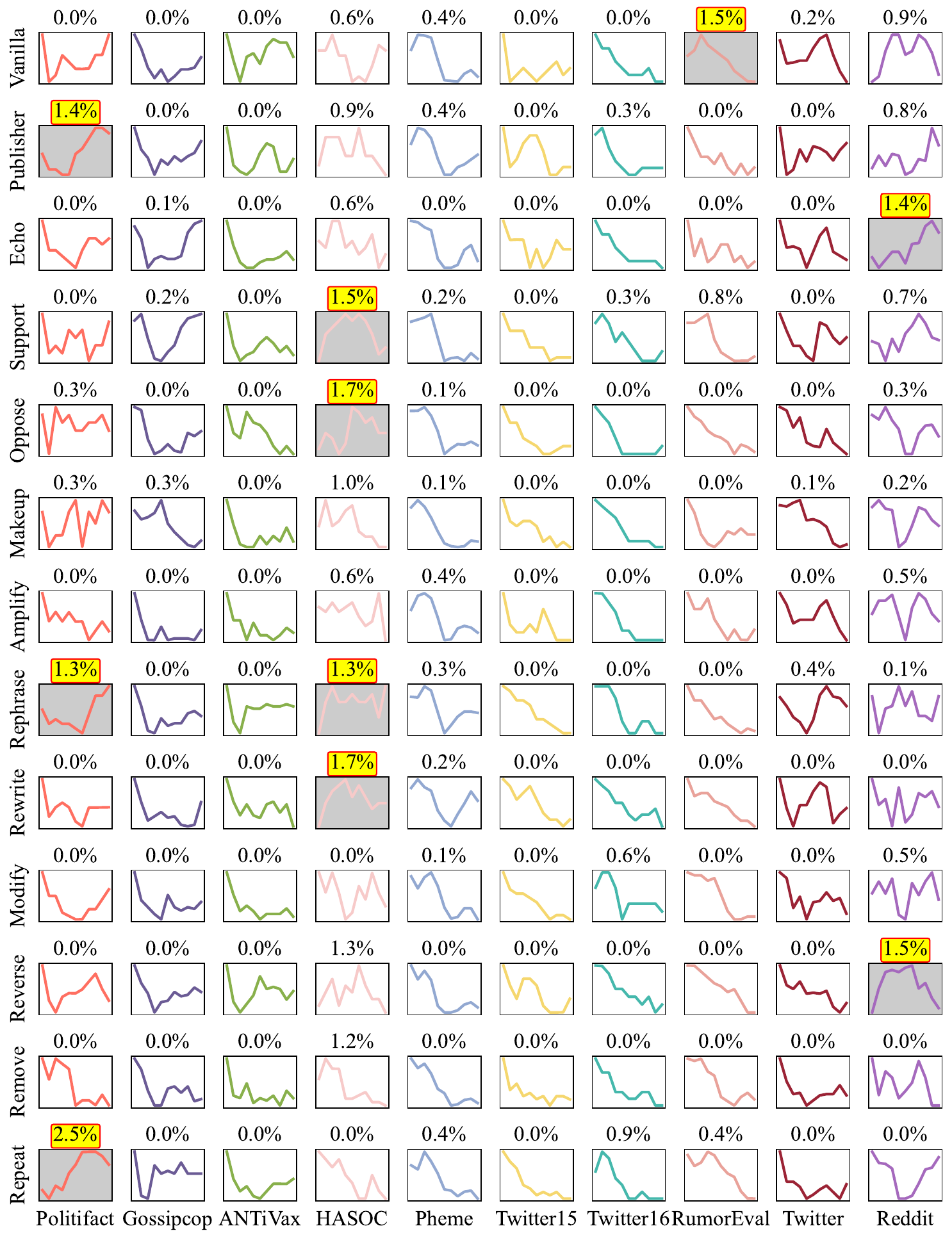}
    \caption{The performance trend of \textbf{Parameter Updating} strategy with re-training data increasing. We present \textsc{GET} on different datasets under different pollution strategies. We present the max improvement of each situation and \colorbox{yellow}{highlight} the top-ten improvement. It strengthens that \textbf{Parameter updating} is the most effective defense strategy, however, the need for annotated data and the unknown when the training ends limit its practical application.}
    \label{fig: updating_get}
\end{figure*}

\begin{figure*}
    \centering
    \includegraphics[width=0.9\linewidth]{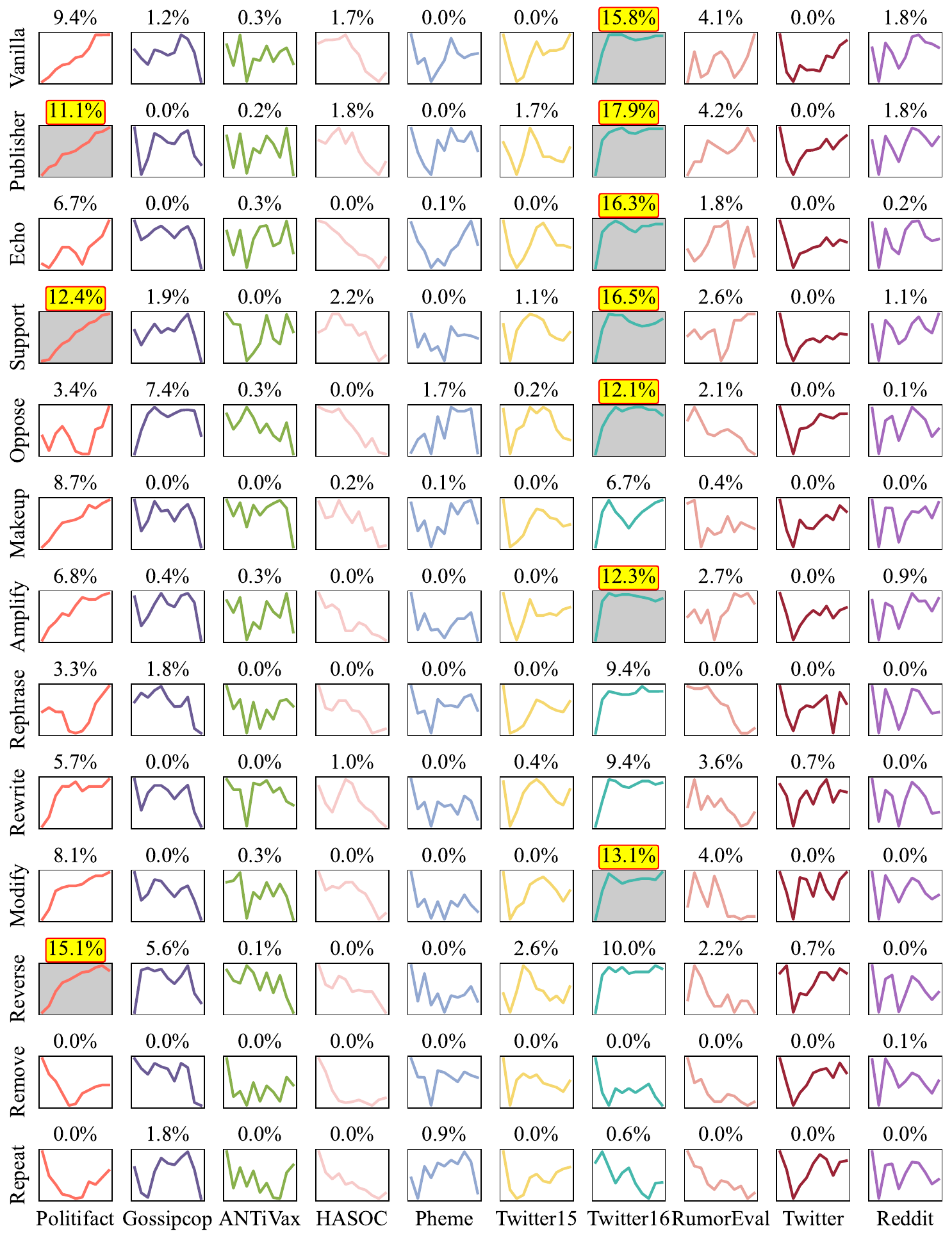}
    \caption{The performance trend of \textbf{Parameter Updating} strategy with re-training data increasing. We present \textsc{BERT} on different datasets under different pollution strategies. We present the max improvement of each situation and \colorbox{yellow}{highlight} the top-ten improvement. It strengthens that \textbf{Parameter updating} is the most effective defense strategy, however, the need for annotated data and the unknown when the training ends limit its practical application.}
    \label{fig: updating_bert}
\end{figure*}

\begin{figure*}
    \centering
    \includegraphics[width=0.95\linewidth]{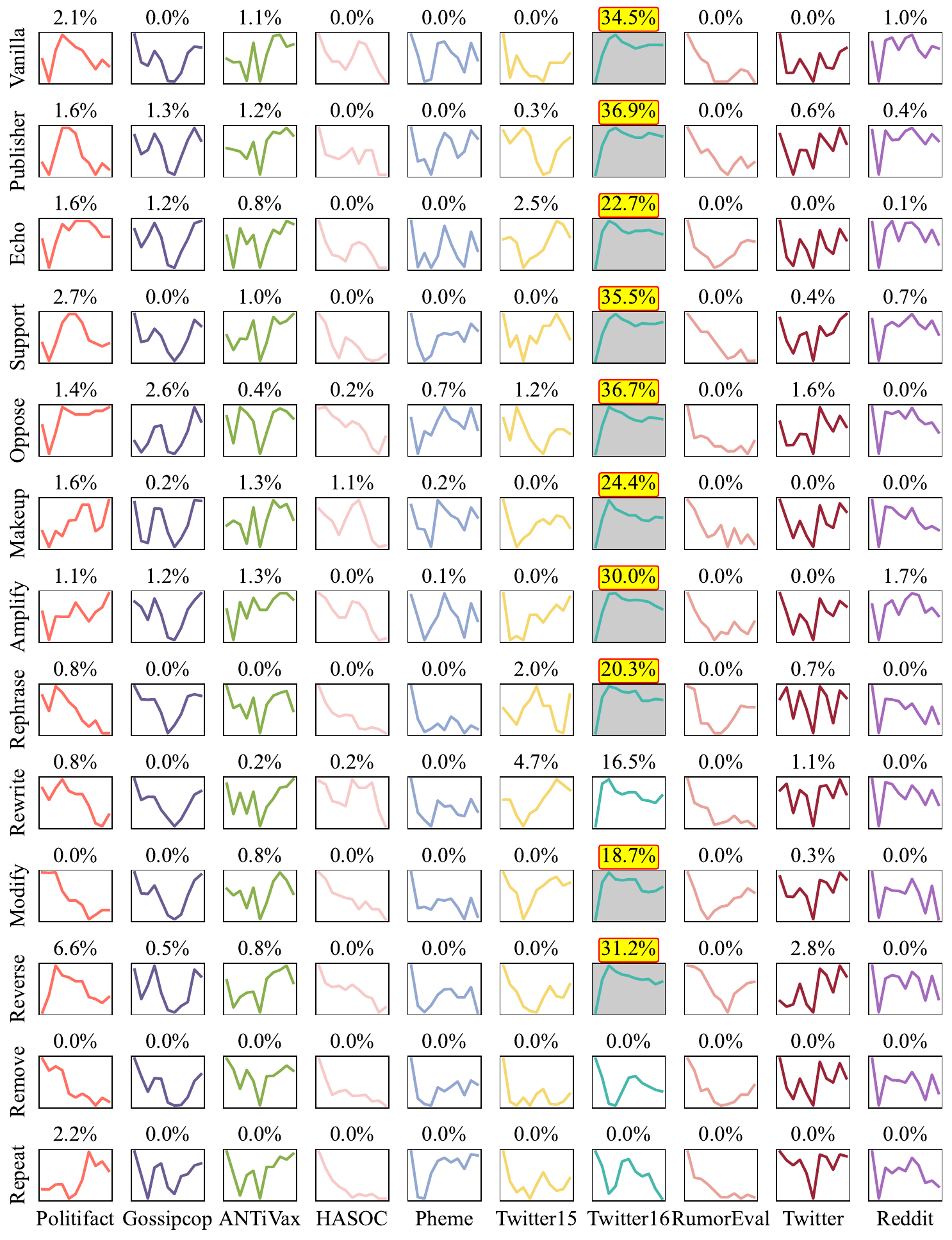}
    \caption{The performance trend of \textbf{Parameter Updating} strategy with re-training data increasing. We present \textsc{DeBERTa} on different datasets under different pollution strategies. We present the max improvement of each situation and \colorbox{yellow}{highlight} the top-ten improvement. It strengthens that \textbf{Parameter updating} is the most effective defense strategy, however, the need for annotated data and the unknown when the training ends limit its practical application.}
    \label{fig: updating_deberta}
\end{figure*}

Figures \ref{fig: updating_defend}, \ref{fig: updating_hyphen}, \ref{fig: updating_get}, \ref{fig: updating_bert}, and \ref{fig: updating_deberta} illustrate the whole results, where we present the improvements and highlight the top-ten performance. The results show that this strategy is the most successful strategy, where the improvements are the most significant. On the other hand, the need for annotated data and the unknown when the training ends limit its practical application.

\section{More Analysis}
\label{app: more_analysis}
\subsection{Human Evaluation}
Among the 29 acceptable annotators, 12 out of 29 prefer generated evidence to original, 14 out of 29 prefer rephrased evidence to original, and 17 out of 29 prefer rephrased evidence to generated.
\subsection{Calibration}
\begin{figure*}
    \centering
    \includegraphics[width=0.85\linewidth]{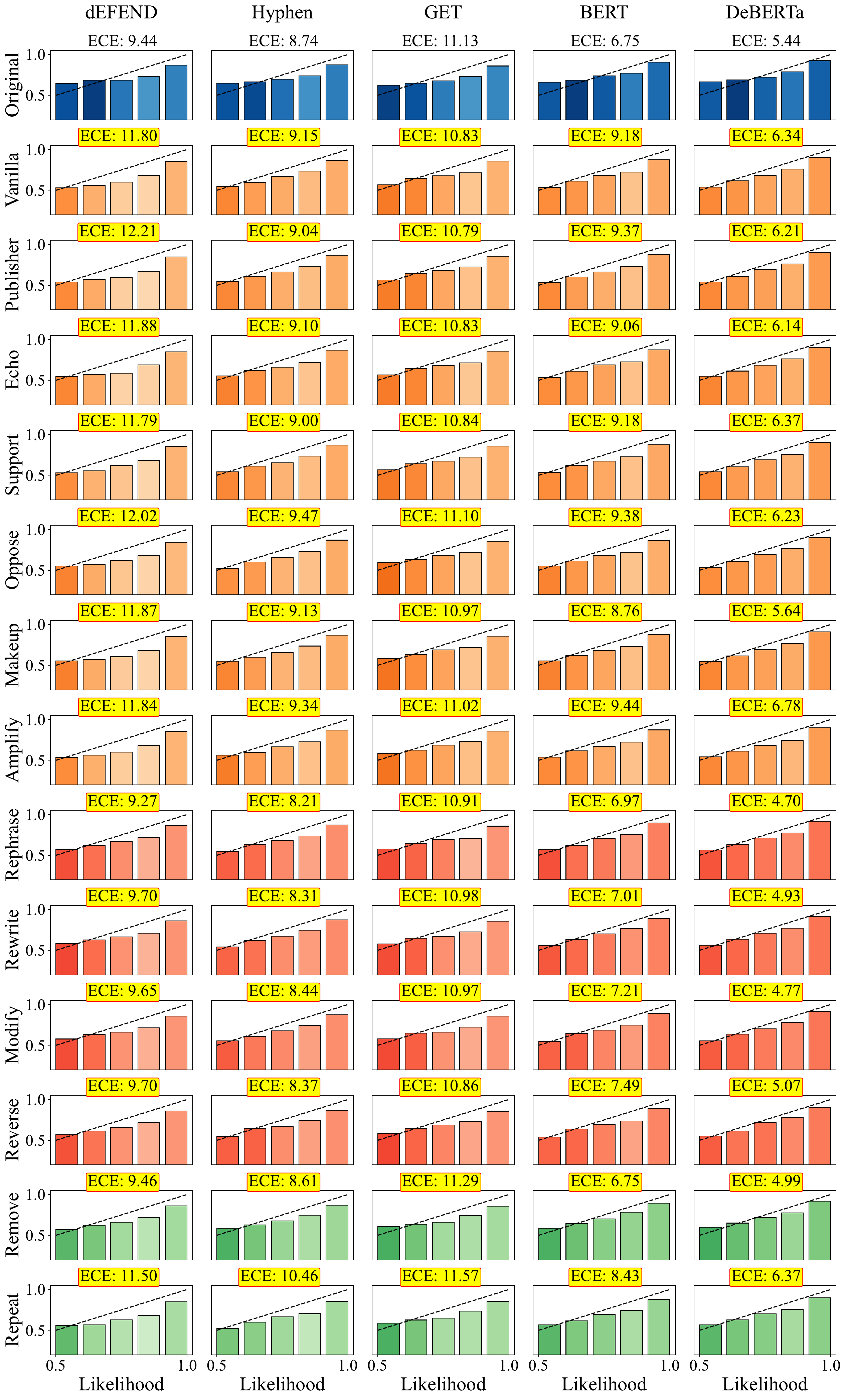}
    \caption{Calibration of existing detectors with the original and polluted evidence. We \colorbox{yellow}{highlight} the values where evidence pollution harms the model calibration. Evidence pollution could harm the model calibration.}
    \label{fig: calibration_appendix}
\end{figure*}

We present the calibration of each baseline under different pollution strategies in Figure \ref{fig: calibration_appendix}. It illustrates that any pollution strategy could harm model calibration.

\begin{table*}[t]
    \centering
    \small
    \begin{tabularx}{\linewidth}{m{0.1\linewidth}|m{0.26\linewidth}|m{0.26\linewidth}|m{0.26\linewidth}}
    \toprule[1.5pt]
        \textbf{Strategy}&\textbf{Content}&\textbf{Original}&\textbf{Polluted}\\
    \midrule[1pt]
    \midrule[1pt]
    \multicolumn{4}{l}{Case Studies for \textbf{Basic Evidence Pollution}.}\\\hline
        \textbf{Remove}&*** may have done irreparable harm to her career this morning when she decided to join a gang of thugs in *** for a day of drinking, drugs and dogfighting at a public park in ***...&This got to be fake news right\newline I truly hope not\newline Who the *** even makes this...\newline *** anyone can create a meme\newline ...his own daughter lol go figure \newline Just like her dad into drugs a thug\newline Hey *** thats a *** story A lie\newline This may be fake news...\newline I LOVE these ** stories...\newline ...realise its not true...&This got to be fake news right\newline I truly hope not\newline Who the *** even makes this...\newline *** anyone can create a meme\newline ...his own daughter lol go figure\\\hline
        \textbf{Repeat}&*** *** Baseball Team To Sell Marijuana Brownies At Their Concession ***, *** –With the new laws in effect in ***, major sports organizations are looking for ways to cash in on the newest legal cash crop in the state...&...are like on level 10 progression\newline Ha?\newline we need to move\newline ...catering to the smokers...\newline I got caught straight slippin’\newline Its very real Ill try to fix the link\newline its fake Heres the link though \newline ...they used unnamed sources...\newline...theyll sell so many snacks\newline food sales are about to blow up&...are like on level 10 progression\newline...are like on level 10 progression\newline...are like on level 10 progression\newline...are like on level 10 progression\newline...are like on level 10 progression\\
    \midrule[1pt]
    \midrule[1pt]
    \multicolumn{4}{l}{Case Studies for \textbf{Rephrase Evidence}.}\\\hline
        \textbf{Rephrase}&\multirow{4}{=}{Attorney General *** today issued the following statement:“During the course of the confirmation proceedings on my nomination to be Attorney General, I advised the Senate Judiciary Committee that ‘[i]f a specific matter arose where I believed my impartiality might reasonably be questioned, I would consult with Department ethics officials regarding the most appropriate way to proceed.’...}&\multirow{4}{=}{do not understand why u \colorbox{green}{hate pres} *** the way u do}&I don't understand why you seem to have such \colorbox{green}{strong dislike} \colorbox{green}{towards President} ***.\\\cline{1-1}\cline{4-4}
        \textbf{Rewrite}&&&I'm not sure why you might think I have a \colorbox{green}{particular opinion} \colorbox{green}{about President} ***. This text is \colorbox{yellow}{simply sharing information} about Attorney General ***'s decision to recuse himself from any investigations related to the 2016 presidential campaigns.\\\cline{1-1}\cline{4-4}
        \textbf{Reverse}&&&It's not clear why you seem to \colorbox{yellow}{support} \colorbox{green}{President} *** so strongly. The Attorney General, ***, has just announced that he will recuse himself from any investigations related to the Presidential campaigns, which suggests a commitment to impartiality and integrity.\\\cline{1-1}\cline{4-4}
        \textbf{Modify}&&&It seems there \colorbox{yellow}{might be} \colorbox{yellow}{a misunderstanding}. This text appears to be a statement from Attorney General *** about recusing himself from any matters related to the 2016 Presidential campaigns. It doesn't \colorbox{green}{express any hate} \colorbox{green}{towards President} ***.\\
    \midrule[1pt]
    \bottomrule[1.5pt]
    \end{tabularx}
    \caption{Case studies of the polluted evidence. We employ the \textbf{Politifact} dataset and conceal personal privacy information such as name and location, as well as content that may contain hate speech. We present the examples and \colorbox{yellow}{highlight} the parts highly related to the intent of a specific pollution strategy. Here are the case studies of \textbf{Basic Evidence Pollution} and \textbf{Rephrase Evidence}. For \textbf{Rephrase Evidence}, we also \colorbox{green}{highlight} the parts sharing the same semantic content as the original. It illustrates that the polluted evidence is of high quality, where LLMs could understand the pollution intent and inject malicious content into the original evidence while keeping it human-like.}
    \label{tab: case_1}
\end{table*}

\begin{table*}[t]
    \centering
    \small
    \begin{tabularx}{\linewidth}{m{0.1\linewidth}|m{0.4\linewidth}|m{0.4\linewidth}}
    \toprule[1.5pt]
        Strategy&Content&Polluted\\
    \midrule[1pt]
    \midrule[1pt]
    \multicolumn{3}{l}{Case Studies for \textbf{Generate Evidence}.}\\\hline
        Vanilla&\multirow{7}{=}{Attorney General *** today issued the following statement:“During the course of the confirmation proceedings on my nomination to be Attorney General, I advised the Senate Judiciary Committee that ‘[i]f a specific matter arose where I believed my impartiality might reasonably be questioned, I would consult with Department ethics officials regarding the most appropriate way to proceed.’...}&*** to step aside from any matters arising from 2016 Presidential campaigns.\\\cline{1-1}\cline{3-3}
        Support&&*** clarified this decision \colorbox{yellow}{doesn't confirm} or suggest the scope of any investigation.\\\cline{1-1}\cline{3-3}
        Oppose&&***'s recusal was a \colorbox{yellow}{tactic, not a commitment}, sources claim\\\cline{1-1}\cline{3-3}
        Publisher&&AG ***'s decision to recuse himself from campaign-related investigations demonstrates commitment to fairness and impartiality. \colorbox{yellow}{\#Transparency}\\\cline{1-1}\cline{3-3}
        Echo&&AG *** has shown his dedication to upholding justice by recusing himself from any matters related to the 2016 Presidential campaigns.\\\cline{1-1}\cline{3-3}
        Makeup&&*** \colorbox{yellow}{consulted with senior career Department} \colorbox{yellow}{officials} before making this decision.\\\cline{1-1}\cline{3-3}
        Amplify&&*** recuses himself from any matters related to 2016 Presidential campaigns. \colorbox{yellow}{What does this mean} \colorbox{yellow}{for the *** investigation? \#Politics}\\
    \midrule[1pt]
    \bottomrule[1.5pt]
    \end{tabularx}
    \caption{Case studies of \textbf{Generate Evidence} (cont.). We present the examples and \colorbox{yellow}{highlight} the parts highly related to the intent of a specific pollution strategy. It illustrates that the generated evidence is of high quality, where LLMs could understand the pollution intent and could inject predetermine malicious content.}
    \label{tab: case_2}
\end{table*}

\end{document}